\documentclass[pmlr,twocolumn,10pt]{jmlr} 

\usepackage{booktabs}
\usepackage{siunitx}

\theorembodyfont{\upshape}
\theoremheaderfont{\scshape}
\theorempostheader{:}
\theoremsep{\newline}

\usepackage{colortbl}
\definecolor{Gray}{gray}{0.85}
\definecolor{aliceblue}{rgb}{0.94, 0.97, 1.0}
\definecolor{beaublue}{rgb}{0.74, 0.83, 0.9}
\definecolor{blond}{rgb}{0.98, 0.94, 0.75}
\definecolor{beige}{rgb}{0.96, 0.96, 0.86}
\definecolor{cornsilk}{rgb}{1.0, 0.97, 0.86}
\definecolor{platinum}{rgb}{0.9, 0.89, 0.89}
\definecolor{blue(pigment)}{rgb}{0.2, 0.2, 0.6}
\definecolor{goldenbrown}{rgb}{0.6, 0.4, 0.08}
\usepackage{enumerate}[leftmargin=*]
\usepackage{enumitem}
\setlist[itemize]{leftmargin=*}
\usepackage{booktabs,ragged2e}
\newcolumntype{Y}{>{\RaggedRight\arraybackslash}X}
\newcolumntype{L}[1]{>{\raggedright\arraybackslash}p{#1}}
\newcolumntype{C}[1]{>{\centering\arraybackslash}p{#1}}
\newcolumntype{M}[1]{>{\centering\arraybackslash}m{#1}}
\newcolumntype{R}[1]{>{\raggedleft\arraybackslash}p{#1}}
\usepackage{graphicx,array}
\usepackage{multirow}
\usepackage{hhline}
\usepackage{amsmath,amssymb,bbm}
\DeclareMathOperator*{\argmin}{arg\,min}

\usepackage{mathtools}

\usepackage{xspace}
\newcommand{\method}{\texttt{AutoMap}\xspace}

\title[Automatic Medical Code Mapping]{\method: Automatic Medical Code Mapping for Clinical Prediction Model Deployment}

\author{
\Name{Zhenbang Wu} \Email{zw12@illinois.edu} \\
\addr{University of Illinois at Urbana-Champaign}
\AND
\Name{Cao Xiao} \Email{danica.xiao@amplitude.com} \\
\addr{Amplitude}
\AND
\Name{Lucas M. Glass} \Email{lucasmglass@gmail.com} \\
\addr{IQVIA}
\AND
\Name{David M. Liebovitz} \Email{david.liebovitz@nm.org} \\
\addr{Northwestern University}
\AND
\Name{Jimeng Sun} \Email{ jimeng.sun@gmail.com} \\
\addr{University of Illinois at Urbana-Champaign}
}

\begin{document}

\maketitle

\begin{abstract}
    Given a deep learning model trained on data from a source site, how to deploy the model to a target hospital automatically? How to accommodate heterogeneous medical coding systems across different hospitals? Standard approaches rely on existing medical code mapping tools, which have significant practical limitations.

To tackle this problem, we propose \method to automatically map the medical codes across different EHR systems in a coarse-to-fine manner: {\bf (1) Ontology-level Alignment:} We leverage the ontology structure to learn a coarse alignment between the source and target medical coding systems; {\bf (2) Code-level Refinement:} We refine the alignment at a fine-grained code level for the downstream tasks using a teacher-student framework.

We evaluate \method using several deep learning models with two real-world EHR datasets: eICU and MIMIC-III. Results show that \method achieves relative improvements up to 3.9\% (AUC-ROC) and 8.7\% (AUC-PR) for mortality prediction, and up to 4.7\% (AUC-ROC) and 3.7\% (F1) for length-of-stay estimation. Further, we show that \method can provide accurate mapping across coding systems. Lastly, we demonstrate that \method can adapt to the two challenging scenarios: (1) mapping between completely different coding systems and (2) between completely different hospitals.
\end{abstract}


\section{Introduction}

Deep learning models have been widely used in clinical predictive modeling with electronic health records (EHRs)~\citep{10.1093/jamia/ocy068}. These models often leverage medical codes as an important data source summarizing patients’ health status~\citep{10.5555/3157382.3157490,10.1145/2939672.2939823,li_behrt_2020}.

However, in real-world clinical practice, a variety of different coding systems are used across hospital EHR systems~\citep{doi:10.1146/annurev-publhealth-031914-122747}. As a result, models trained on data from a source hospital are often hard to adapt to a target hospital where other coding systems are used. A method that can {\bf accommodate different medical coding systems across hospitals for easy model deployment} is highly desirable. Standard approaches rely on existing medical code mapping tools (e.g., Unified Medical Language System (UMLS)~\citep{pmid14681409}), which have significant practical limitations due to the following challenges:
\begin{itemize}[leftmargin=*]
    \item \textbf{Rare coding systems:} Existing commercial and free code mapping tools are only available for a few widely used coding systems (e.g., among ICD-9, ICD-10 and SNOMED CT), as creating such tools requires substantial human efforts~\citep{wojcik_challenge_2006}. Hospitals using some rare or even private coding systems cannot benefit from the mapping tools.
    \item \textbf{Limited labeled data:} While large hospitals may fine-tune the pre-trained models to adapt to their coding systems, small hospitals with limited labeled data often fail to do so.
    \item \textbf{No access to source data:} Worse still, the source data usually cannot be shared with the target hospital due to privacy and legal concern.
\end{itemize}

In this paper, we propose \method for automatic medical code mapping across different hospitals EHR systems. \method constructs appropriate target embeddings unsupervisedly based on the target EHR data and maps the target embeddings to the source embeddings, so that the deep learning model trained on the source data can be seamlessly deployed to the target data without any manual code mapping. More specifically, \method learns the mapping across different coding systems in a coarse-to-fine manner:
\begin{itemize}[leftmargin=*]
\item {\bf Embedding.} The medical code embeddings will be constructed from the target EHR data unsupervisedly.
\item {\bf Ontology-level Alignment.} We leverage the ontology structure to map medical coding groups via iterative self-supervised learning. In this step, we obtain a coarse mapping matrix from groups of target embeddings to the groups of source embeddings.
\item {\bf Code-level Refinement.} We refine the mapping matrix at a fine-grained code level via a  teacher-student framework. It utilizes a discriminator (teacher A) to align two coding systems at the code level, and the backbone model (teacher B) to optimize the mapping based on the final prediction.
\end{itemize}

We evaluate \method using multiple backbone deep learning models and test with two real-world EHR datasets: eICU~\citep{pollard_eicu_2018} and MIMIC-III~\citep{mimiciii}.

Results show that with a limited set of labeled data, \method achieves relative improvements up to $3.9\%$ on AUC-ROC score and $8.7\%$ on AUC-PR score for mortality prediction; and up to $4.7\%$ on AUC-ROC score and $3.7\%$ on F1 score for length-of-stay estimation. Further, we evaluate the mapping accuracy of \method and show that \method improves the best baseline method by $8.2\%$ in similarity score and $11.3\%$ on hit@10 score. Lastly, we demonstrate that \method can still achieve acceptable results under the two challenging scenarios: (1) mapping between completely different coding systems: the model is trained on diagnosis codes and deployed on medication codes; (2) mapping between completely different hospitals: the model is trained and deployed in hospitals from different regions.
\section{Related Work}

\noindent\textbf{EHR Representation Learning.}
Deep learning models have been widely used in EHR representation learning especially for modeling discrete medical codes~\citep{10.1093/jamia/ocy068}. Existing models were either designed to capture complex and temporal patterns in EHR data~\citep{10.5555/3157382.3157490,pmlr-v56-Choi16,10.1145/2939672.2939823,7762861,10.1145/3097983.3097997,10.1145/3097983.3098088,Harutyunyan_2019,DBLP:conf/aaai/MaGWZWRTGM20,Ma_Zhang_Wang_Ruan_Wang_Tang_Ma_Gao_Gao_2020,10.1145/3394486.3403107,Zhang_Gao_Ma_Wang_Wang_Tang_2021,DBLP:conf/aaai/XuSD21}, model structural information in medical codes~\citep{10.1145/3097983.3098126,DBLP:conf/aaai/ShangXMLS19,10.1609/aaai.v34i01.5400}, or augment the model using pre-training~\citep{li_behrt_2020,PMID:34017034,STEINBERG2021103637} and memory network~\citep{DBLP:conf/aaai/ShangXMLS19}. However, most of the existing works focus on the model building phase while ignoring the challenge of model deployment due to the diversity of EHR coding formats~\citep{doi:10.1146/annurev-publhealth-031914-122747}.\\

\noindent\textbf{Medical Code Mapping Tools.}
There exists a variety of commercial and free tools for mapping across different EHR ecosystems. UMLS~\citep{pmid14681409} provides the mapping among ICD-9, ICD-10 and SNOMED CT. Observational Medical Outcomes Partnership (OMOP)~\citep{pmid26262116} and Fast Healthcare Interoperability Resources (FHIR)~\citep{pmid26911829} define the standards for representing clinical data in a consistent format. Relying on these tools, some recent works try to support model deployment across hospitals by transforming the EHR data into a standard format~\citep{rajkomar_scalable_2018,tang_democratizing_2020}. However, creating such tools requires a lot of domain knowledge and human labor~\citep{wojcik_challenge_2006}. These mapping tools are only available for widely-used coding systems and easily outdated due to code updates. To our best knowledge, \method is the first work towards code-agnostic deep learning deployment in the healthcare domain without relying on existing code mapping tools. \\

\noindent\textbf{Transfer Learning.}
Transfer learning focuses on improving a target task using the knowledge obtained from a model trained for a related source task~\citep{NIPS1992_67e103b0}. \cite{pmlr-v56-Choi16} empirically confirm that RNN models possess great potential for transfer learning across different hospitals. \cite{gupta2019transfer} transfer the knowledge with parameter sharing using a deep RNN such that the target task only needs to fine-tune a simpler linear classifier. However, both works assume consistent data format between source and target data. Recently, \cite{ma2020covidcare} distill knowledge from  EHR data to enhance the prognosis for inpatients with emerging infectious diseases. A separate RNN is used for each feature to improve the compatibility across datasets with different feature sets. However, \cite{ma2020covidcare} still require the availability of both source and target datasets, and a subset of shared features between them. Thus, it does not work in our setting since we only have access to the target dataset with no shared features.\\

\noindent\textbf{Cross-lingual Word Mapping.}
Our medical code mapping problem has some similarity to the cross-lingual research. Cross-lingual word mapping methods work by mapping the word embeddings in two languages to a shared space using translation pairs~\citep{artetxe-etal-2017-learning}, shared tokens~\citep{sogaard-etal-2018-limitations}, adversarial learning~\citep{conneau2018word}, or the nearest neighbors of similarity distributions~\citep{artetxe-etal-2018-robust}.

Inspired by \cite{artetxe-etal-2018-robust}, \method also leverages the similarity distributions~\citep{mikolov2013exploiting} to align medical codes. However, there are significant differences between EHR and natural languages: (1)  medical codes often reside in a concept hierarchy; (2) medical codes are often noisier. To address this, instead of directly mapping medical codes, \method adopts a coarse-to-fine method by first performing ontology-level alignment and then code-level refinement.

\section{Method}

\begin{figure*}[th]
\begin{center}
\includegraphics[width=0.8\linewidth]{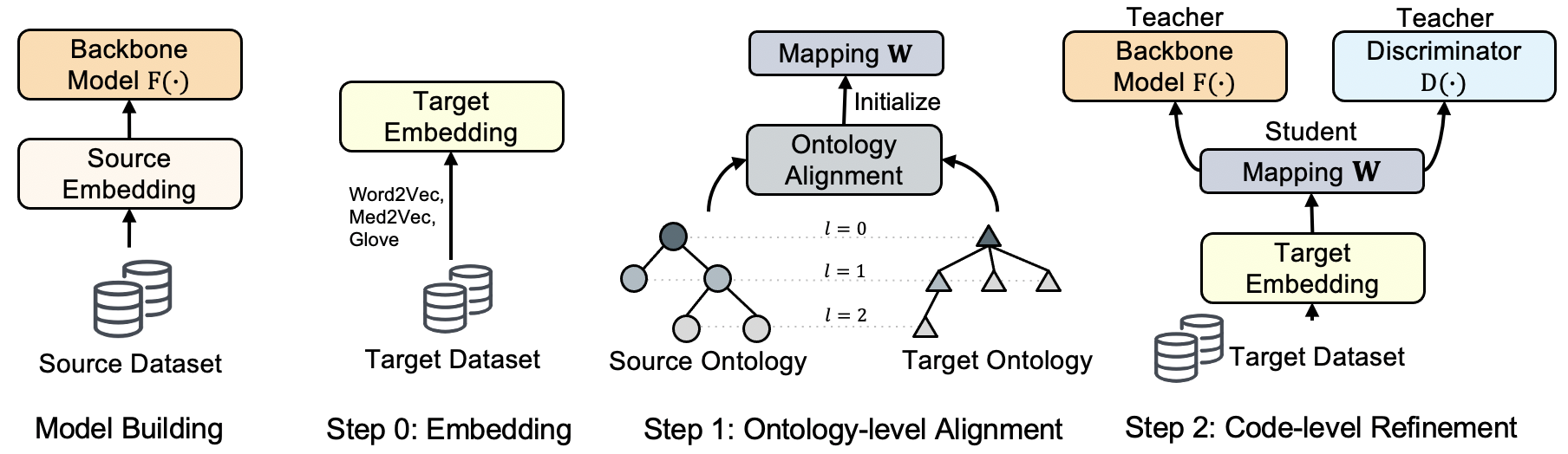}
\end{center}
\caption{An overview of \method. 
\method supports model deployment by automatically mapping the medical code embeddings across different coding systems in a coarse-to-fine manner: (0) Embedding that initializes the target code embedding matrix; (1) Ontology-level Alignment that leverages the ontology structure to learn the coarse ontology mapping; (2) Code-level Refinement that refines the mapping at the fine-grained code level for the downstream task with a teacher-student framework.} 
\label{fig:model}
\end{figure*}

\subsection{Problem Setting}
We first define a few key concepts, and then present our setting in Prob.~\ref{prob: setting}. Detailed notations can be found in Appx.~\ref{appendix:notations}.

\begin{definition}[\textbf{EHR Dataset}]
In the EHR data, a patient has a sequence of visits: $V_p = [ {v}_p^{(1)} , {v}_p^{(2)} , \dots , {v}_p^{(n_p)} ]$, where $n_p$ is the number of visits of patient $p$. For model training, each patient has a label $\mathbf{y}_p$ (e.g., mortality or length-of-stay). We will drop the subscript $_{p}$ whenever it is unambiguous. Each visit of a patient is represented by its corresponding medical codes, specified by ${v}^{(i)} = \{ \mathbf{c}_1, \mathbf{c}_2, \dots, \mathbf{c}_{m^{(i)}} \}$, where $m^{(i)}$ is the total number of codes of the $i$-th visit. Each medical code $\mathbf{c} \in \{0, 1\}^{|\mathcal{C}|}$ is a one-hot vector (i.e., $\Vert \mathbf{c} \Vert_1 = 1$), where $\mathcal{C}$ denotes the set of all medical codes in the dataset.
\end{definition}

Our setting involves two datasets: a source dataset $*S$ for pre-training the backbone model but unavailable during deployment, and a mostly unlabeled target dataset $*T$ for deploying the model. The two datasets can have completely different medical codes. We also utilize separate medical ontology structures for source and target medical codes.

\begin{definition}[\textbf{Medical Ontology}]
A medical ontology $\mathcal{O}$  specifies the hierarchy of medical codes in the form of a parent-child relationship. Formally, an ontology $\mathcal{O}$ is a directed acyclic graph whose nodes are $\mathcal{C} \cup \overline{\mathcal{C}}$. Here, $\mathcal{C}$ is the set of medical codes (often leaf nodes in the ontology), and $\overline{\mathcal{C}}$ is the set of other intermediate codes (i.e., non-leaf nodes) representing more general concepts.
\end{definition}

For simplicity, we deﬁne a function $\text{ancestor}(\mathbf{c}, l): \{0, 1\}^{|\mathcal{C}|} \times \mathbb{Z} \rightarrow \{0, 1\}^{|\overline{\mathcal{C}}|}$, which maps a given medical code $\mathbf{c} \in \{0, 1\}^{|\mathcal{C}|}$ to its $l$-th level ancestor code (i.e., category). For example, in Fig.~\ref{fig:model}, the root node is the $0$-th level ancestor code of all leaf codes.

\begin{definition}[\textbf{Medical Code Embedding}]
\label{def: emb}
To fully utilize the code semantic information, it is a common practice to convert the medical code from one-hot vector $\mathbf{c} \in \{0, 1\}^{|\mathcal{C}|}$ to a dense embedding vector $\mathbf{e} \in \mathbb{R}^{d}$~\citep{10.5555/3157382.3157490,li_behrt_2020}, where $d$ is the embedding dimensionality. This can be done via an embedding matrix $\mathbf{E} \in \mathbb{R}^{|\mathcal{C}| \times d}$, where each row corresponds to the embedding for a medical code. The embedding can be computed as $\mathbf{e} = \mathbf{E}^\top\mathbf{c}$.
\end{definition}

We denote the embedding matrices for source and target datasets as $\mathbf{E}_S$ and $\mathbf{E}_T$, respectively. The source embedding $\mathbf{E}_S$ is provided with the trained backbone model as the input. And the target embedding $\mathbf{E}_T$ will be learned using the target dataset. In this work, to deploy the backbone model, we will learn a mapping from the source to the target embedding space.

\begin{definition}[\textbf{Code Embedding Mapping}]
We define the mapping from the embedding space of one medical coding system to another as $\phi(\mathbf{E}): \mathbb{R}^d \rightarrow \mathbb{R}^d$.
\end{definition}

We will learn the embedding mapping $\phi(\cdot)$ that maps the target embedding to the source via $\phi(\mathbf{E}_T)$.

\begin{definition}[\textbf{Backbone Deep Learning Model}]
The backbone deep learning model $F(\cdot)$ takes EHR sequences and the corresponding medical code embeddings as the input and then outputs the prediction: $\hat{\mathbf{y}} = F([{v}^{(i)}]_{i=1}^{n}, \phi(\mathbf{E}))$, where $\hat{\mathbf{y}}$ is the corresponding predictions for label $\mathbf{y}$. The backbone model $F(\cdot)$ is pre-trained on source dataset $*S$ and deployed on  target dataset $*T$ with a different coding system. Note that the embedding mapping $\phi(\cdot)$ degenerates to the identity function if the backbone model $F(\cdot)$ is trained and deployed on the same coding system.
\end{definition}

We evaluate \method with different backbone models ranging from classic MLP and RNN to more recent GNN and attention-based models on the following tasks.

\begin{task}[\textbf{Mortality Prediction}]
Given a patient's EHR data $[{v}^{(i)}]_{i=1}^{n}$, we aim to predict mortality $y_M \in \{0, 1\}$ using the backbone model $F_M(\cdot)$. Formally, $\hat{y}_M = F_M([{v}^{(i)}]_{i=1}^{n}, \phi(\mathbf{E}))$,
where $\hat{y}_M \in \{0, 1\}$. 
\end{task}

\begin{task}[\textbf{Length-of-Stay Estimation}]
Given a patient's EHR data $[{v}^{(i)}]_{i=1}^{n}$, we aim to predict the patient's length of stay $\mathbf{y}_L \in \mathbb{R}^{n}$ using a backbone model $F_R(\cdot)$. We follow previous works~\citep{Harutyunyan_2019} to categorize label $\mathbf{y}_L$ into four classes $\{0, 1, 2, 3\}$, i.e., less than 1 day (class 0), 1 to 7 days (class 1), 7 to 14 days (class 2), and more than 14 days (class 3). In this way, we frame length-of-stay estimation as a multi-class classification task. Formally, $\hat{\mathbf{y}}_L = F_R([{v}^{(i)}]_{i=1}^{n}, \phi(\mathbf{E}))$, where $\hat{\mathbf{y}}_L \in \{0, 1, 2, 3\}^{n}$. 
\end{task}

We note that \method is general and can support other prediction tasks as well.

\begin{problem}[\textbf{Predictive Model Deployment}]
\label{prob: setting}
Given a backbone model $F(\cdot)$ and source code embedding matrix $\mathbf{E}_S$, a mostly unlabeled target dataset $*T$ in a different coding system, and the medical ontologies $\mathcal{O}_S, \mathcal{O}_T$ for both coding systems, the goal is to optimize the mapping 
$\phi(\cdot)$ on the target dataset $*T$, as given by Eq.~\eqref{eq:opt},
\begin{equation}
\argmin_{\phi(\cdot)} \mathcal{L}(F(\cdot), \mathbf{E}_S, *T, \mathcal{O}_S, \mathcal{O}_T, \phi(\cdot)),
\label{eq:opt}
\end{equation}
where $\mathcal{L}(\cdot)$ denotes the designated loss function. The prediction on the target dataset can be obtained via $F(V_T, \phi(\mathbf{E}_T))$, where $V_T$ is a sequence of visits from the target dataset $*T$, and $\phi(\mathbf{E}_T)$ is the transformed target embeddings.
\end{problem}

In our setting, we can only access the source code embedding $\mathbf{E}_S$ and ontology $\mathcal{O}_S$ instead of the source data $*S$. This is more realistic in deployment setting since the source data often cannot be shared due to legal and privacy concern. In contrast, the source embedding matrix $\mathbf{E}_S$ can be more easily provided along with the backbone model $F(\cdot)$, and the code ontologies are usually publicly accessible. We also assume that the target dataset $*T$ is mostly unlabeled, since the target site may often be some small hospital.

\subsection{The \method Method}
We propose \method for automatic code mapping across different hospitals EHR systems. The mapping will be done in a coarse-to-fine manner, enabled by the adaptation process shown in Fig.~\ref{fig:model}. Embedding (step 0) first initializes the target code embedding matrix $\mathbf{E}_T$. Ontology-level alignment (step 1) then derives the initial coarse mapping $\phi(\cdot)$ via iterative self-supervised learning. Code-level refinement (step 2) further fine-tunes the mapping $\phi(\cdot)$ at the code level using a teacher-student framework.

\subsubsection{Step 0: Embedding.}
As mentioned in Def.~\ref{def: emb}, we first convert the target medical codes from one-hot vector $\mathbf{c}_T \in \{0, 1\}^{|\mathcal{C}|}$ to a corresponding dense embedding vector $\mathbf{e}_T \in \mathbb{R}^{d}$. We use GloVe~\citep{pennington2014glove} to learn the target code embedding matrix $\mathbf{E}_T$ via a global co-occurrence matrix of medical codes. Other unsupervised learning algorithms such as Med2Vec~\citep{10.1145/2939672.2939823} and Word2Vec~\citep{mikolov2013efficient} can also be used. We employ GloVe because of its computational efficiency.

Code embeddings can also be computed based on code text descriptions. However, the text descriptions may bring in additional noise, which may confuse the mapping. Moreover, under the extreme case where the model is trained and deployed on two completely different coding systems (e.g., from diagnosis codes to medication codes), the text descriptions may not help since the descriptions are too different. On the contrary, we show empirically that \method can still adapt to this extreme case.

After that, we parameterize $\phi(\cdot)$ by a mapping matrix $\mathbf{W} \in \mathbb{R}^{d \times d}$. The mapping matrix $\mathbf{W}$ can be used to transform the target code embedding via $\mathbf{E}_T\mathbf{W}$.

\subsubsection{Step 1: Ontology-level Alignment.}
In this step, we will first learn a coarse mapping $\mathbf{W}$ at the ontology level. This first step is essential because direct code level mapping is difficult and unnecessary: (1) It is difficult due to the large number of  medical codes; (2) It is also unnecessary since many codes have similar clinical meanings. Therefore, we follow a common practice to first group similar codes using code ontology~\citep{10.1145/2939672.2939823,10.5555/3157382.3157490,DBLP:conf/aaai/ShangXMLS19} and learn the mapping on groups instead of leaf-level codes.
For example, ICD-9 code 438.11 ``late effects of cerebrovascular disease, aphasia'' corresponds to five ICD-10 codes (I69.020, I69.120, I69.220, I69.320, I69.920). While it is hard to directly align the ICD-9 code to each of these five ICD-10 codes, we can first coarsely map the ICD-9 code to I00-I99 ``diseases of the circulatory system'', and then gradually refine the mapping to I60-I99 ``cerebrovascular diseases'', I69 ``cerebrovascular diseases'', and eventually the five-leaf codes. By leveraging the medical ontology, we can use more general medical concepts as ``anchor points'' to better align two coding systems. 

Next, we introduce the building blocks of the iterative self-supervised learning  (i.e., ontology grouping, unsupervised seed induction, Procrustes refinement), and then present the ontology-level alignment algorithm.\\

\noindent\textbf{Ontology Grouping.} At a given hierarchy level $l$, we group the codes according to their $l$-th level ontology categories. 
Specifically, the $i$-th group $\mathcal{G}_i^{(l)}$ consists of all the leaf medical codes whose $l$-th level category is $\mathbf{c}_i$, as in Eq.~\eqref{eq:group},
\begin{equation}
\label{eq:group}
\mathcal{G}_i^{(l)} = \{ \mathbf{c}_j \mid \text{ancestor}(\mathbf{c}_j, l) = \mathbf{c}_i, \mathbf{c}_j \in \mathcal{C} \},
\end{equation}
where $\mathbf{c}_i \in \overline{\mathcal{C}}$ is the corresponding $l$-th level category code. We will drop the superscript $^{(l)}$ whenever it is unambiguous. 

To obtain the group embedding $\mathbf{g}_i$, we first calculate the mean group embedding $\overline{\mathbf{g}}_i$ by averaging all the code embeddings in that group, as in Eq.~\eqref{eq:group_mean_emb}; then, we represent the group embedding as the closest code embedding, as in Eq.~\eqref{eq:group_med_emb},
\begin{subequations}
\label{eq:group_emb}
\begin{align}
\overline{\mathbf{g}}_i &= \text{mean}\{ \mathbf{e}_j \mid \mathbf{c}_j \in \mathcal{G}_i \} \label{eq:group_mean_emb}, \\
\mathbf{g}_i &= \underset{\mathbf{e}_j}{\text{argmin}} \{ \mathbf{e}_j \overline{\mathbf{g}}_i^\top \mid  \mathbf{c}_j \in \mathcal{C} \} \label{eq:group_med_emb},
\end{align} 
\end{subequations}
where $\mathbf{e}_j$ is the embedding vector for the code $\mathbf{c}_j$, and $\mathbf{e}_j \overline{\mathbf{g}}_i^\top \in \mathbb{R}$ calculates the distance between the code $\mathbf{c}_j$ and the mean group embedding $\overline{\mathbf{g}}_i$. Intuitively, $\mathbf{g}_i$ can be viewed as the ``median'' group embedding. We select the top-$k$ groups based on the group size, since we want to first learn a coarse mapping, while including too many groups may introduce too much granular information. As a result, we have $\mathbf{G}_T, \mathbf{G}_S  \in \mathbb{R}^{k \times d}$ for target and source groups, where each row corresponds to an embedding vector for a particular group. We present with the same $k$ to reduce clutter, though it can be different for source and target groups.

We note that when the ontology is not available, \method can still apply by using a clustering algorithm (e.g., k-Means) to group the medical codes. Specifically, we provide additional experiments on this setting in Appx.~\ref{appendix:kmeans}.\\

\noindent\textbf{Unsupervised Seed Induction.} Given the $l$-th level source and target coding groups $\mathbf{G}_S$ and $\mathbf{G}_T$, we can initialize a coarse alignment in a fully unsupervised way. More specifically, we first calculate the similarity matrices, as in Eq.~\eqref{eq:similarity_matrices},
\begin{equation}
\label{eq:similarity_matrices}
\mathbf{M}_T = \mathbf{G}_T^{}\mathbf{G}_T^\top; \; \mathbf{M}_S = \mathbf{G}_S^{}\mathbf{G}_S^\top,
\end{equation}
where $\mathbf{M}_T, \mathbf{M}_S \in \mathbb{R}^{k \times k}$. Each row in the similarity matrices $\mathbf{M}_T, \mathbf{M}_S$ represents the similarities of the corresponding group to all the other groups. Under the ideal case where the embedding spaces between different coding systems are isometric\footnote{In practice, the isometry requirement will not hold exactly, but it can be assumed to hold approximately, or the problem of mapping two code embedding spaces without supervision would be impossible.}, one can permute the rows and columns of $\mathbf{M}_T$ to obtain $\mathbf{M}_S$. We introduce the following heuristics to find the optimal permutation (i.e., a mapping dictionary) of this NP-hard problem~\citep{Karp1972}. We perform row-wise sort on $\mathbf{M}_T$ and $\mathbf{M}_S$ (i.e., elements in each row are sorted based only on the order in that particular row), as in Eq.~\eqref{eq:sort}. Under the isometric assumption, codes with the same meaning will have exactly the same row vector in $\tilde{\mathbf{M}}_T$ and $\tilde{\mathbf{M}}_S$, suggesting that we can find the mapping dictionary $\mathbf{D} \in \mathbb{R}^{k \times k}$ via nearest neighbor search over row vectors in $\tilde{\mathbf{M}}_T$ , as shown in Eq.~\eqref{eq:init_d},
\begin{subequations}
\label{eq:init}
\begin{align}
\tilde{\mathbf{M}}_{T} &= \text{sorted}(\mathbf{M}_{T}); \tilde{\mathbf{M}}_{S} = \text{sorted}(\mathbf{M}_{S}), \label{eq:sort} \\
\mathbf{D}[i,j] &= 
\begin{cases}
1 ,& \text{if} \; j = \text{argmax}( (\tilde{\mathbf{M}}_{T}^{} \cdot \tilde{\mathbf{M}}_{S}^\top)[i, :] ) \label{eq:init_d} \\
0 ,& \text{otherwise},
\end{cases}
\end{align}
\end{subequations}
where $\cdot$ denotes matrix multiplication.\\

\noindent\textbf{Procrustes Optimization.} At a given hierarchy level $l$, we optimize the inducted mapping dictionary $\mathbf{D}$ by iterating the following two steps.

\begin{enumerate}[leftmargin=*]
    \item The mapping $\mathbf{W} \in \mathbb{R}^{d \times d}$ is obtained by maximizing the similarities for the current dictionary $\mathbf{D}$ as given by Eq.~\eqref{eq:procrustes}. This optimization problem is known as the Procrustes problem~\citep{schonemann_generalized_1966} and has a closed form solution, as in Eq.~\eqref{eq:optimal_w},
    \begin{subequations}
    \label{eq:update_w}
    \begin{align}
    & \underset{\mathbf{W}}{\text{argmin}} \Vert \mathbf{D} \odot ( \underbrace{\mathbf{G}_{T} \mathbf{W}}_{\mathclap{\text{transformed target embedding}}} \mathbf{G}_{S}^\top ) \Vert_1 \label{eq:procrustes}, \\
    \mathbf{W} = \mathbf{U}\mathbf{V}^\top & \text{, where  } \mathbf{U}\mathbf{\Sigma}\mathbf{V}^T = \text{SVD}(\mathbf{G}_T^{\top}\mathbf{D}\mathbf{G}_S^{}) \label{eq:optimal_w},
    \end{align}
    \end{subequations}
    where $\odot$ denotes Hadamard product, and SVD denotes Singular Value Decomposition~\citep{1102314}. 
    
    \item A new dictionary $\mathbf{D}$ is induced, as in Eq.~\eqref{eq:update_d},
    \begin{equation}
    \setlength{\belowdisplayskip}{\baselineskip}
    \label{eq:update_d}
    \mathbf{D}[i,j] = 
    \begin{cases}
    1 ,& \text{if} \; j = \text{argmax}( (\mathbf{G}_{T} \mathbf{W} \mathbf{G}_{S}^\top)[i, :] ) \\
    0 ,& \text{otherwise.}
    \end{cases}
    \end{equation}
\end{enumerate}

\noindent\textbf{Iterative Self-supervised Learning.} We now introduce the self-supervised learning strategy, which maps the two coding systems at multiple resolutions iteratively. Starting from a coarse hierarchy level $l$, we obtain the $l$-th level medical coding groups $\mathbf{G}_S$ and $\mathbf{G}_T$ with Eq.~(\ref{eq:group},~\ref{eq:group_emb}). Then we induct the $l$-th level seed mapping dictionary $\mathbf{D}^{(l)}$ with Eq.~(\ref{eq:similarity_matrices},~\ref{eq:init}). Next, we merge the current and previous level mapping dictionaries, as $\mathbf{D}^{(l)} = \mathbf{D}^{(l)} + \mathbf{D}^{(l-1)}$. Lastly, we optimize the merged mapping dictionary $\mathbf{D}^{(l)}$ using Eq.~(\ref{eq:update_w},~\ref{eq:update_d}). We gradually increase $l$ (going down in the ontology) during iterative self-supervised learning until we reach the leaf level to learn the mapping at multiple resolutions. We note that source and target codes can use different grouping level $l$. We present with the same $l$ to reduce clutter.

In this way, we learn a coarse mapping matrix $\mathbf{W}$ between two medical coding systems at the ontology level. This step is inspired by \cite{artetxe-etal-2018-robust}. However, instead of directly mapping medical codes, \method leverages the ontology structure and iteratively maps medical coding groups in a coarse-to-fine manner, allowing \method to better align coding systems with different granularities.

\subsubsection{Step 2: Code-level Refinement.}
While we have performed step 1 (ontology-level alignment) to initialize the mapping, the mapping is still too coarse and need further refining. Moreover, there is no guarantee of the performance for the downstream tasks (i.e., mortality prediction and length-of-stay estimation). Thus, it is preferred to further fine-tune the mapping at the code level for downstream tasks.

To do this, we propose a teacher-student framework, where the discriminator $D(\cdot)$ (teacher A) refines the mapping matrix $\mathbf{W}$ (student) via adversarial learning; and the backbone model $F(\cdot)$ (teacher B) optimizes the mapping matrix $\mathbf{W}$ (student) based on the final prediction task. Below we describe the framework in  detail.\\

\noindent\textbf{Teacher A: Discriminator.}
We leverage the adversarial learning framework by introducing a discriminator $D(\cdot)$, parameterized by a multi-layer neural network. Specifically, the discriminator $D(\cdot)$ tries to classify whether the embeddings are from the target (label 0) or source (label 1) embedding distributions. Formally, discriminator $D(\cdot)$ aims at minimizing the discriminator adversarial loss, as in Eq.~\eqref{eq:loss_d},
\begin{equation}
\label{eq:loss_d}
\mathcal{L}_D = -\log( D(\mathbf{e}_S) ) -\log( D(1 - \mathbf{e}_T\mathbf{W}) ),
\end{equation}
where $\mathbf{e}_S$ ($\mathbf{e}_T$) represents the source (target) code embedding sampled randomly from the code embedding matrix $\mathbf{E}_S$ ($\mathbf{E}_T$), and $\mathbf{W}$ maps the target embedding to the source embedding space via $\mathbf{e}_T\mathbf{W}$.

The mapping matrix $\mathbf{W}$ acts as the generator and tries to deceive the discriminator $D(\cdot)$. Formally, we try to minimize the generator adversarial loss, as in Eq.~\ref{eq:loss_g},
\begin{equation}
\label{eq:loss_g}
\mathcal{L}_G = -\log( D(\mathbf{e}_T\mathbf{W}) ).
\end{equation}
Theoretically, the discriminator $D(\cdot)$ and mapping matrix $\mathbf{W}$ learn to align two coding systems as an adversarial game, which is essentially minimizing the following Jensen-Shannon divergence (JSD)~\citep{goodfellow2014generative}, as shown in Eq.~\eqref{eq.jsd}, 
\begin{equation}
\text{JSD}(p(\mathbf{e}_S) \Vert p(\mathbf{e}_T\mathbf{W})).
\label{eq.jsd}
\end{equation}
The definition of JSD can be found in Appx.~\ref{appendix:divergence}. Note that since the minimization happens at the distribution level, we do not require code mapping pairs to supervise training.\\

\noindent\textbf{Teacher B: Backbone.} 
Here, the backbone model $F(\cdot)$ is leveraged to optimize the ultimate prediction performance based on the transformed target code embeddings. Formally, we aim at minimizing the following classification loss
\begin{equation}
\label{eq:loss_cls}
\mathcal{L}_\text{cls}(F([{v}^{(i)}]_{i=1}^{n}, \mathbf{E}_T\mathbf{W}), \mathbf{y}_T),
\end{equation}
where the transformed target code embeddings $\mathbf{E}_T\mathbf{W}$ are used to encode patient visits $[{v}^{(i)}]_{i=1}^{n}$.

In summary, the mapping matrix $\mathbf{W}$ is  fine-tuned by minimizing the  combined loss 
\begin{equation}
\label{eq:loss_w}
\mathcal{L}_W = \mathcal{L}_\text{cls} + \alpha \mathcal{L}_G,
\end{equation}
where $\alpha$ is a hyper-parameter to balance the two teachers. 
The pseudo-code of \method can be found in Appx.~\ref{sec:code}.

\subsubsection{Remarks.}  
Step 2 (code-level refinement) assumes a white-box backbone model $F(\cdot)$. If $F(\cdot)$ is a blackbox model, we will perform step 1 (ontology-level alignment).

\section{Experiment}

We deploy \method with several backbone deep learning models and test on two real-world EHR datasets to answer the following questions:
\begin{enumerate}[leftmargin=*]
\item Can \method deploy backbone deep learning models to target hospitals with limited labeled data?
\item Can \method learn accurate mapping across different coding systems?
\item Can \method adapt to target hospitals with completely different coding systems?
\item Can \method adapt to target hospitals from completely different regions?
\end{enumerate}

\subsection{Experimental Setting}
\noindent\textbf{Data.} We evaluate the performance of \method extensively with two publicly accessible datasets: eICU~\citep{pollard_eicu_2018} and MIMIC-III~\citep{mimiciii}. eICU~\citep{pollard_eicu_2018} is a multi-center database with intensive care unit (ICU) records for over 200K admissions to over 200 hospitals across the United States. MIMIC-III~\citep{mimiciii} is a single-center database containing 53K ICU records from Beth Israel Deaconess Medical Center.

For Q1 (limited labeled data) and Q2 (mapping accuracy), we use eICU~\citep{pollard_eicu_2018} to evaluate the model performance across ICD-9 and ICD-10 codes. For Q3 (different coding systems), we evaluate this scenario with MIMIC-III~\citep{mimiciii}. The backbone model is trained on diagnosis codes (ICD-9) and then deployed on medication codes (NDC). For Q4 (different hospitals), we use the multi-center database eICU~\citep{pollard_eicu_2018}. Following previous work~\citep{10.1145/3450439.3451878}, we train the backbone model in hospitals from Midwest region and deploy it to hospitals from South region. Detailed setting and statistics can be found in Appx.~\ref{appendix:dataset}.\\

\noindent\textbf{Baselines.} Next, we present all the baselines.
\begin{itemize}[leftmargin=*]
    \item {\bf Direct Training:} Directly train a new model on the target dataset.
    \item {\bf Transfer Learning:} Swap the embedding matrix and fine-tune the new embedding with the backbone model.
    \item {\bf Mapping Tools:} Map target medical codes to the source using existing code mapping tools.
    \item {\bf MUSE~\citep{conneau2018word}:} Learn the mapping by aligning the target and source embedding distributions via adversarial learning.
    \item {\bf VecMap~\citep{artetxe-etal-2018-robust}:} Learn the mapping via iterative self-supervised learning based on the code-level structure similarity.
\end{itemize}
Lastly, we conduct an ablation study of our \method.
\begin{itemize}
    \item {\bf Step 1 Only:} We only perform step 1 (ontology-level alignment) to learn the mapping matrix $\mathbf{W}$. Then we directly use the transformed target embedding for the downstream task. This is to evaluate the contribution of the ontology mapping.
    \item {\bf Step 1 Only + Random Ontology:} We only perform step 1 (ontology-level alignment) but with a randomly generated ontology. This is to evaluate the contribution of leveraging medical ontology.
    \item {\bf Step 2 Only:} We randomly initialize the mapping matrix $\mathbf{W}$ and then only perform step 2 code-level refinement. This is to evaluate the contribution of the teacher-student framework.
\end{itemize}

\noindent\textbf{Backbone Models.} As \method is a general framework that can apply to different backbone models, we incorporate \method with the following backbone deep learning models:
\begin{itemize}[leftmargin=*]
    \item {\bf MLP:} Learn the visit representation using a simple feed-forward neural network.
    \item {\bf RNN:} Learn the visit representation using a simple recurrent neural network.
    \item {\bf RETAIN~\citep{10.5555/3157382.3157490}:} A two-level neural  attention model which can detect influential past visits and significant clinical variable within those visits.
    \item {\bf GCT~\citep{10.1609/aaai.v34i01.5400}:} A graph convolutional transformer which jointly learns the hidden structure of EHR while performing supervised prediction tasks on EHR data.
    \item {\bf BEHRT~\citep{li_behrt_2020}:} A deep neural sequence transduction model based on BERT~\citep{devlin-etal-2019-bert}, which performs self-attention mechanism~\citep{vaswani2017attention} on the sequence of visits. 
\end{itemize}

\subsection{Q1: Target Data with Limited Labels}
\begin{table}[h]
\begin{center}
\resizebox{\linewidth}{!}{
\begin{tabular}{l|l|c|c}
\toprule
\multirow{2}{*}{\bf Backbone} & \multirow{2}{*}{\bf Method} &  {\bf Mortality} & {\bf Length-of-Stay} \\
& & AUC-PR & F1 \\
\midrule
\multirow{6}{*}{MLP} & {\it Full-Label} & {\it 0.2819 $\pm$ 0.0317} & {\it 0.5033 $\pm$ 0.0169} \\
\cline{2-4}
& Direct Training & 0.2524 $\pm$ 0.0285 & 0.2835 $\pm$ 0.0147 \\
& Transfer Learning & 0.2551 $\pm$ 0.0287 & 0.4584 $\pm$ 0.0179 \\
& MUSE & 0.2506 $\pm$ 0.0279 & 0.4905 $\pm$ 0.0172 \\
& VecMap & 0.2820 $\pm$ 0.0315 & 0.4947 $\pm$ 0.0171 \\
\rowcolor{platinum}
\cellcolor{white} & \method &  {\bf 0.2934 $\pm$ 0.0324*} & {\bf 0.4952 $\pm$ 0.0168} \\
\midrule
\multirow{6}{*}{RNN} & {\it Full-Label} & {\it 0.2818 $\pm$ 0.0319} & {\it 0.5030 $\pm$ 0.0167} \\
\cline{2-4}
& Direct Training & 0.2074 $\pm$ 0.0236 & 0.1222 $\pm$ 0.0089 \\
& Transfer Learning & 0.2536 $\pm$ 0.0286 & 0.4662 $\pm$ 0.0176 \\
& MUSE  & 0.2455 $\pm$ 0.0275 & 0.4933 $\pm$ 0.0169 \\
& VecMap & 0.2780 $\pm$ 0.0311 & {\bf 0.5019 $\pm$ 0.0168} \\
\rowcolor{platinum}
\cellcolor{white} & \method & {\bf 0.2875 $\pm$ 0.0319*} & {0.4996 $\pm$ 0.0163} \\
\midrule
\multirow{6}{*}{RETAIN} & {\it Full-Label} & {\it 0.2648 $\pm$ 0.0302} & {\it 0.4447 $\pm$ 0.0183} \\
\cline{2-4}
& Direct Training & 0.2031 $\pm$ 0.0228 & 0.1222 $\pm$ 0.0089 \\
& Transfer Learning & 0.2269 $\pm$ 0.0262 & 0.4455 $\pm$ 0.0179 \\
& MUSE & 0.2374 $\pm$ 0.0283 & 0.4217 $\pm$ 0.0185 \\
& VecMap & 0.2744 $\pm$ 0.0305 & 0.4264 $\pm$ 0.0184 \\
\rowcolor{platinum}
\cellcolor{white} & \method & {\bf 0.2835 $\pm$ 0.0313*} & {\bf 0.4779 $\pm$ 0.0167*} \\
\midrule
\multirow{6}{*}{GCT} & {\it Full-Label} & {\it 0.2814 $\pm$ 0.0323} & {\it 0.4986 $\pm$ 0.0163} \\
\cline{2-4}
& Direct Training & 0.1836 $\pm$ 0.0189 & 0.2680 $\pm$ 0.0158 \\
& Transfer Learning & 0.2103 $\pm$ 0.0222 & 0.4748 $\pm$ 0.0162 \\
& MUSE & 0.2242 $\pm$ 0.0250 & 0.4866 $\pm$ 0.0159 \\
& VecMap & 0.2491 $\pm$ 0.0264 & 0.4863 $\pm$ 0.0162 \\
\rowcolor{platinum}
\cellcolor{white} & \method & {\bf 0.2707 $\pm$ 0.0294*} & {\bf 0.4940 $\pm$ 0.0164*} \\
\midrule
\multirow{6}{*}{BEHRT} & {\it Full-Label} & {\it 0.2652 $\pm$ 0.0275} & {\it 0.3657 $\pm$ 0.0176} \\
\cline{2-4}
& Direct Training & 0.1740 $\pm$ 0.0163 & 0.3063 $\pm$ 0.0163 \\
& Transfer Learning & 0.2320 $\pm$ 0.0249 & 0.3291 $\pm$ 0.0178 \\
& MUSE & 0.2155 $\pm$ 0.0222 & 0.3493 $\pm$ 0.0178 \\
& VecMap & {\bf 0.2786 $\pm$ 0.0292} & 0.3612 $\pm$ 0.0179 \\
\rowcolor{platinum}
\cellcolor{white} & \method & {0.2712 $\pm$ 0.0280} & {\bf 0.3744 $\pm$ 0.0182*} \\
\bottomrule
\end{tabular}
}
\caption{Results with limited labeled data ($100$ patients) in the target site. Dataset is eICU~\citep{pollard_eicu_2018}. The average scores of two mapping directions between ICD-9 and ICD-10 codes are reported. $\pm$ denotes standard deviations. * indicates that \method achieves significant improvement over the best baseline method (i.e., p-value is smaller than $0.05$). Experiment results show that \method can adapt different backbone models to the target site with limited labeled data.}
\label{tab:limited_label}
\end{center}
\end{table}

We first evaluate \method in a common setting where the target site has limited labeled data ($100$ patients). Results can be found in in Tab.~\ref{tab:limited_label}. For reference, we also report the performance of the model trained with the fully-labeled target data, as {\it ``Full-Label''} in the table. This can be viewed as an ``upper bound'' of the model performance. Due to the limited space, we only report AUC-PR for mortality and F1 for length-of-stay here. Additional results can be found in Appx.~\ref{appendix:q1_cont}. Descriptions of the metrics can be found in Appx~\ref{appendix:metrics}.

As shown in Tab.~\ref{tab:limited_label}, first, we find that the two simple baselines: direct training and transfer learning methods do not work very well. In most cases, they are much worse compared to the full-label performance. This is expected as the amount of labeled data is insufficient to train or fine-tune the backbone models. Next, code-level mapping methods MUSE~\citep{conneau2018word} and VecMap~\citep{artetxe-etal-2018-robust} achieve some improvements, but they are not stable. In some cases, they perform even worse than the two simple baselines. This may because ICD-9 and ICD-10 have different degrees of specificity (e.g., 10K codes in ICD-9 v.s. 68K codes in ICD-10), and directly mapping them at code level does not work very well. Finally, we observe that \method achieves significant improvement over the baseline and can match the full-label performance in most cases. Specifically, \method achieves up to $8.7\%$ relative improvement on AUC-PR score for mortality prediction; for length-of-stay estimation, \method achieves up to and $3.7\%$ relative improvement on F1 score. This demonstrates the effectiveness of coarse-to-fine mapping and the versatility of \method.

\subsection{Q2: Mapping Accuracy}
\begin{table}[h]
\begin{center}
\resizebox{\linewidth}{!}{
\begin{tabular}{lcc}
\toprule
{\bf Method} & {\bf Similarity} & {\bf Hit@10} \\
\midrule
MUSE  & 0.1633 $\pm$ 0.0110 & 0.0600 $\pm$ 0.0113 \\
VecMap & 0.4612 $\pm$ 0.0980 & 0.5974 $\pm$ 0.1841 \\
\rowcolor{platinum}
\method & {\bf 0.4992 $\pm$ 0.0070*} & {\bf 0.6657 $\pm$ 0.0187*}\\
\bottomrule
\end{tabular}
}
\caption{Accuracy of mapping for diagnosis codes (ICD-9 and ICD-10). Dataset is eICU~\citep{pollard_eicu_2018}. The average scores of two mapping directions are reported. Experiment results show that \method can learn accurate mapping across medical coding systems.}
\label{tab:code_mapping}
\end{center}
\end{table}

We then evaluate the accuracy of the learnt mapping. The ICD code mapping in the eICU~\citep{pollard_eicu_2018} dataset is used as the ground truth.  Due to the limited space, for the rest of the experiments, we only report results with BEHRT~\citep{li_behrt_2020} using $100$ labeled patients in the target data.

As shown in Tab.~\ref{tab:code_mapping}, VecMap~\citep{artetxe-etal-2018-robust} and \method achieve much better performance than MUSE~\citep{conneau2018word}. This supports the isometric assumption used in both methods. Further, \method achieves the best results with statistical significance. This demonstrates that the proposed coarse-to-fine mapping can better map coding systems with different granularities.

\subsection{Q3: Completely Different Codes}
\begin{table}[h!]
\begin{center}
\resizebox{\columnwidth}{!}{
\begin{tabular}{lcc}
\toprule
\multirow{2}{*}{\bf Method} &  {\bf Mortality} & {\bf Length-of-Stay} \\
& AUC-PR & F1 \\
\midrule
{\it Full-Label} & {\it 0.7149 $\pm$ 0.0227} & {\it 0.3057 $\pm$ 0.0173} \\
\hline
Direct Training & 0.4701 $\pm$ 0.0236 & {\bf 0.3158 $\pm$ 0.0174} \\
Transfer Learning & 0.5642 $\pm$ 0.0255 & 0.2999 $\pm$ 0.0171 \\
MUSE & 0.4905 $\pm$ 0.0241 & 0.3022 $\pm$ 0.0171 \\
VecMap & 0.3553 $\pm$ 0.0173 & 0.3014 $\pm$ 0.0171 \\
\rowcolor{platinum}
\method & {\bf 0.5902 $\pm$ 0.0252*} & {0.3022 $\pm$ 0.0171} \\
\bottomrule
\end{tabular}}
\caption{Results for the scenario where the backbone model is trained on diagnosis code (ICD-9) and deployed on medication codes (NDC). Dataset is MIMIC-III~\citep{mimiciii}. Experiment results show that \method can adapt to target data coded in a completely different system.}
\label{tab:diag_to_med}
\end{center}
\end{table}

We next evaluate \method on the challenging case where we train the backbone model on diagnosis code (ICD-9) and deploy it on medication codes (NDC). Results can be found in in Tab.~\ref{tab:diag_to_med}.

First, we note that since these two coding systems are so different, no existing mapping tools is available. For mortality prediction, as shown in Tab.~\ref{tab:diag_to_med}, the code-level mapping methods perform even worse than direct training and transfer learning. This may due to the large gap between these two coding systems. On the contrary, \method can still give acceptable results, outperforming all baseline methods with $4.6\% - 66.1\%$ statistically significant improvements. This shows the superiority of \method's coarse-to-fine mapping strategy. For the length-of-stay estimation task, all five methods perform pretty similar to full-label performance. This may indicate that medication codes are not so informational for length-of-stay estimation.

\subsection{Q4: Completely Different Hospitals}
\begin{table}[h!]
\begin{center}
\resizebox{\columnwidth}{!}{
\begin{tabular}{lcc}
\toprule
\multirow{2}{*}{\bf Method} &  {\bf Mortality} & {\bf Length-of-Stay} \\
& AUC-PR & F1 \\
\midrule
{\it Full-Label} & {\it 0.2578 $\pm$ 0.0300} & {\it 0.4560 $\pm$ 0.0158} \\
\hline
Direct Training & 0.1434 $\pm$ 0.0154 & {\bf 0.4334 $\pm$ 0.0164} \\
Transfer Learning & 0.1860 $\pm$ 0.0214 & 0.3924 $\pm$ 0.0158 \\
MUSE & 0.1314 $\pm$ 0.0150 & 0.3988 $\pm$ 0.0165 \\
VecMap & 0.1305 $\pm$ 0.0163 & 0.3801 $\pm$ 0.0167 \\
\rowcolor{platinum}
\method & {\bf 0.1990 $\pm$ 0.022*} & {0.4290 $\pm$ 0.0157} \\
\bottomrule
\end{tabular}}
\caption{Results for the scenario where the backbone model is trained and deployed in hospitals from different regions. Dataset is eICU~\citep{pollard_eicu_2018}. Experiment results show that \method can adapt to target hospitals from a completely region.}
\label{tab:diff_region}
\end{center}
\end{table}

We further challenge \method under the scenario where we train the backbone model in hospitals from Midwest region (with ICD-9 code) and deploy it in hospitals from South region (with ICD-10 code). Results can be found in in Tab.~\ref{tab:diff_region}.

For mortality prediction, mapping based methods (MUSE~\citep{conneau2018word} and VecMap~\citep{artetxe-etal-2018-robust}) achieve the worst results. This is expected as methods from cross-lingual word mapping do not consider the domain gap between different regions. This also explains why transfer learning perform slightly better (as its training scheme can accommodate some domain gap). Benefit from the refinement step, \method achieves the best result with $7.0\% - 52.5\%$ statistically significant relative improvements. This shows that \method can adapt to hospitals from different regions. For length-of-stay estimation, all pre-training based methods perform worse than direct training. This may indicate that different hospitals have different decision rules on ICU length-of-stay. As a result, transferring knowledge from other hospitals may not help. Despite this, \method still achieves the best results among all pre-training based methods.

\subsection{Ablation study}
\begin{table}[h!]
\begin{center}
\resizebox{\columnwidth}{!}{
\begin{tabular}{lcc}
\toprule
\multirow{2}{*}{\bf Method} &  {\bf Mortality} & {\bf Length-of-Stay} \\
& AUC-PR & F1 \\
\midrule
Step 1 Only & 0.2680 $\pm$ 0.0275 & 0.3623 $\pm$ 0.0181 \\
Step 1 Only + R.O. & 0.2054 $\pm$ 0.0215 & 0.3631 $\pm$ 0.0179 \\
Step 2 Only & 0.2038 $\pm$ 0.0213 & 0.3306 $\pm$ 0.0181 \\
\rowcolor{platinum}
\method & {\bf 0.2712 $\pm$ 0.0280*} & {\bf 0.3744 $\pm$ 0.0182*} \\
\bottomrule
\end{tabular}}
\caption{Ablation study. Dataset is eICU~\citep{pollard_eicu_2018}. The average scores of two mapping directions between ICD-9 and ICD-10 codes are reported. R.O. denotes random ontology. Experiment results demonstrate the importance of \method's 2-step coarse-to-fine mapping.}
\label{tab:ablation}
\end{center}
\end{table}

Finally, we compare \method with three ablated versions. As shown in Tab.~\ref{tab:ablation}, only performing step 2 (code-level refinement) gives the worst results. This is reasonable as the model will easily over-fit the target data with limited labels. Also, since the mapping matrix $\mathbf{W}$ is randomly generated, the adversarial learning module will even harm the downstream tasks. Next, we can see that performing step 1 (ontology-level alignment) only gives better results. This indicates that step 1 contributes most to \method's improvements. This may because the isometric assumption and medical ontology can act as a strong prior to guide the model learning process. This point can also be supported by the performance with randomly-generated ontology. Lastly, \method achieves the best results. This shows the importance of refining the mapping at code-level after the coarse ontology alignment.
\section{Conclusion}

We propose \method, an automatic for automatic medical code mapping across different hospitals EHR systems. \method is enabled by the following process: (0) Embedding: We construct target embeddings from a target dataset. (1) Ontology-level Alignment: We leverage the ontology structure to learn a coarse ontology-level mapping; (2) Code-level Refinement: We further refine the mapping at the fine-grained code level for the downstream task. Benefit from this coarse-to-fine mapping, \method can better align coding systems at different granularities. We evaluate \method extensively using different backbone models with two real-world EHR datasets. Experimental results show that \method outperforms existing solutions on multiple prediction tasks when mapping solutions exist and provides a mapping strategy when conventional solutions do not exist. To our best knowledge, \method is the first work towards code-agnostic deep learning deployment in the healthcare domain without relying on existing code mapping tools.

\section*{Institutional Review Board (IRB)}
This research does not require IRB approval.


\bibliography{jmlr-sample}

\begin{thebibliography}{48}
\providecommand{\natexlab}[1]{#1}
\providecommand{\url}[1]{\texttt{#1}}
\expandafter\ifx\csname urlstyle\endcsname\relax
  \providecommand{\doi}[1]{doi: #1}\else
  \providecommand{\doi}{doi: \begingroup \urlstyle{rm}\Url}\fi

\bibitem[Artetxe et~al.(2017)Artetxe, Labaka, and
  Agirre]{artetxe-etal-2017-learning}
Mikel Artetxe, Gorka Labaka, and Eneko Agirre.
\newblock Learning bilingual word embeddings with (almost) no bilingual data.
\newblock In \emph{Proceedings of the 55th Annual Meeting of the Association
  for Computational Linguistics (Volume 1: Long Papers)}, pages 451--462,
  Vancouver, Canada, July 2017. Association for Computational Linguistics.
\newblock \doi{10.18653/v1/P17-1042}.
\newblock URL \url{https://www.aclweb.org/anthology/P17-1042}.

\bibitem[Artetxe et~al.(2018)Artetxe, Labaka, and
  Agirre]{artetxe-etal-2018-robust}
Mikel Artetxe, Gorka Labaka, and Eneko Agirre.
\newblock A robust self-learning method for fully unsupervised cross-lingual
  mappings of word embeddings.
\newblock In \emph{Proceedings of the 56th Annual Meeting of the Association
  for Computational Linguistics (Volume 1: Long Papers)}, pages 789--798,
  Melbourne, Australia, July 2018. Association for Computational Linguistics.
\newblock \doi{10.18653/v1/P18-1073}.
\newblock URL \url{https://www.aclweb.org/anthology/P18-1073}.

\bibitem[Baytas et~al.(2017)Baytas, Xiao, Zhang, Wang, Jain, and
  Zhou]{10.1145/3097983.3097997}
Inci~M. Baytas, Cao Xiao, Xi~Zhang, Fei Wang, Anil~K. Jain, and Jiayu Zhou.
\newblock Patient subtyping via time-aware lstm networks.
\newblock In \emph{Proceedings of the 23rd ACM SIGKDD International Conference
  on Knowledge Discovery and Data Mining}, KDD '17, page 65–74, New York, NY,
  USA, 2017. Association for Computing Machinery.
\newblock ISBN 9781450348874.
\newblock \doi{10.1145/3097983.3097997}.
\newblock URL \url{https://doi.org/10.1145/3097983.3097997}.

\bibitem[Birkhead et~al.(2015)Birkhead, Klompas, and
  Shah]{doi:10.1146/annurev-publhealth-031914-122747}
Guthrie~S. Birkhead, Michael Klompas, and Nirav~R. Shah.
\newblock Uses of electronic health records for public health surveillance to
  advance public health.
\newblock \emph{Annual Review of Public Health}, 36\penalty0 (1):\penalty0
  345--359, 2015.
\newblock \doi{10.1146/annurev-publhealth-031914-122747}.
\newblock URL \url{https://doi.org/10.1146/annurev-publhealth-031914-122747}.
\newblock PMID: 25581157.

\bibitem[Bodenreider(2004)]{pmid14681409}
O.~Bodenreider.
\newblock {{T}he {U}nified {M}edical {L}anguage {S}ystem ({U}{M}{L}{S}):
  integrating biomedical terminology}.
\newblock \emph{Nucleic Acids Res}, 32\penalty0 (Database issue):\penalty0
  D267--270, Jan 2004.

\bibitem[Choi et~al.(2016{\natexlab{a}})Choi, Bahadori, Kulas, Schuetz,
  Stewart, and Sun]{10.5555/3157382.3157490}
Edward Choi, Mohammad~Taha Bahadori, Joshua~A. Kulas, Andy Schuetz, Walter~F.
  Stewart, and Jimeng Sun.
\newblock Retain: An interpretable predictive model for healthcare using
  reverse time attention mechanism.
\newblock In \emph{Proceedings of the 30th International Conference on Neural
  Information Processing Systems}, NIPS'16, page 3512–3520, Red Hook, NY,
  USA, 2016{\natexlab{a}}. Curran Associates Inc.
\newblock ISBN 9781510838819.

\bibitem[Choi et~al.(2016{\natexlab{b}})Choi, Bahadori, Schuetz, Stewart, and
  Sun]{pmlr-v56-Choi16}
Edward Choi, Mohammad~Taha Bahadori, Andy Schuetz, Walter~F. Stewart, and
  Jimeng Sun.
\newblock Doctor ai: Predicting clinical events via recurrent neural networks.
\newblock In Finale Doshi-Velez, Jim Fackler, David Kale, Byron Wallace, and
  Jenna Wiens, editors, \emph{Proceedings of the 1st Machine Learning for
  Healthcare Conference}, volume~56 of \emph{Proceedings of Machine Learning
  Research}, pages 301--318, Northeastern University, Boston, MA, USA, 18--19
  Aug 2016{\natexlab{b}}. PMLR.
\newblock URL \url{https://proceedings.mlr.press/v56/Choi16.html}.

\bibitem[Choi et~al.(2016{\natexlab{c}})Choi, Bahadori, Searles, Coffey,
  Thompson, Bost, Tejedor-Sojo, and Sun]{10.1145/2939672.2939823}
Edward Choi, Mohammad~Taha Bahadori, Elizabeth Searles, Catherine Coffey,
  Michael Thompson, James Bost, Javier Tejedor-Sojo, and Jimeng Sun.
\newblock Multi-layer representation learning for medical concepts.
\newblock In \emph{Proceedings of the 22nd ACM SIGKDD International Conference
  on Knowledge Discovery and Data Mining}, KDD '16, page 1495–1504, New York,
  NY, USA, 2016{\natexlab{c}}. Association for Computing Machinery.
\newblock ISBN 9781450342322.
\newblock \doi{10.1145/2939672.2939823}.
\newblock URL \url{https://doi.org/10.1145/2939672.2939823}.

\bibitem[Choi et~al.(2017)Choi, Bahadori, Song, Stewart, and
  Sun]{10.1145/3097983.3098126}
Edward Choi, Mohammad~Taha Bahadori, Le~Song, Walter~F. Stewart, and Jimeng
  Sun.
\newblock Gram: Graph-based attention model for healthcare representation
  learning.
\newblock In \emph{Proceedings of the 23rd ACM SIGKDD International Conference
  on Knowledge Discovery and Data Mining}, KDD '17, page 787–795, New York,
  NY, USA, 2017. Association for Computing Machinery.
\newblock ISBN 9781450348874.
\newblock \doi{10.1145/3097983.3098126}.
\newblock URL \url{https://doi.org/10.1145/3097983.3098126}.

\bibitem[Choi et~al.(2020)Choi, Xu, Li, Dusenberry, Flores, Xue, and
  Dai]{10.1609/aaai.v34i01.5400}
Edward Choi, Zhen Xu, Yujia Li, Michael Dusenberry, Gerardo Flores, Emily Xue,
  and Andrew Dai.
\newblock Learning the graphical structure of electronic health records with
  graph convolutional transformer.
\newblock \emph{Proceedings of the AAAI Conference on Artificial Intelligence},
  34:\penalty0 606--613, 04 2020.
\newblock \doi{10.1609/aaai.v34i01.5400}.

\bibitem[Conneau et~al.(2018)Conneau, Lample, Ranzato, Denoyer, and
  Jégou]{conneau2018word}
Alexis Conneau, Guillaume Lample, Marc'Aurelio Ranzato, Ludovic Denoyer, and
  Hervé Jégou.
\newblock Word translation without parallel data, 2018.

\bibitem[Devlin et~al.(2019)Devlin, Chang, Lee, and
  Toutanova]{devlin-etal-2019-bert}
Jacob Devlin, Ming-Wei Chang, Kenton Lee, and Kristina Toutanova.
\newblock {BERT}: Pre-training of deep bidirectional transformers for language
  understanding.
\newblock In \emph{Proceedings of the 2019 Conference of the North {A}merican
  Chapter of the Association for Computational Linguistics: Human Language
  Technologies, Volume 1 (Long and Short Papers)}, pages 4171--4186,
  Minneapolis, Minnesota, June 2019. Association for Computational Linguistics.
\newblock \doi{10.18653/v1/N19-1423}.
\newblock URL \url{https://www.aclweb.org/anthology/N19-1423}.

\bibitem[Goodfellow et~al.(2014)Goodfellow, Pouget-Abadie, Mirza, Xu,
  Warde-Farley, Ozair, Courville, and Bengio]{goodfellow2014generative}
Ian~J. Goodfellow, Jean Pouget-Abadie, Mehdi Mirza, Bing Xu, David
  Warde-Farley, Sherjil Ozair, Aaron Courville, and Yoshua Bengio.
\newblock Generative adversarial networks, 2014.

\bibitem[Gupta et~al.(2019)Gupta, Malhotra, Narwariya, Vig, and
  Shroff]{gupta2019transfer}
Priyanka Gupta, Pankaj Malhotra, Jyoti Narwariya, Lovekesh Vig, and Gautam
  Shroff.
\newblock Transfer learning for clinical time series analysis using deep neural
  networks, 2019.

\bibitem[Harutyunyan et~al.(2019)Harutyunyan, Khachatrian, Kale, Ver~Steeg, and
  Galstyan]{Harutyunyan_2019}
Hrayr Harutyunyan, Hrant Khachatrian, David~C. Kale, Greg Ver~Steeg, and Aram
  Galstyan.
\newblock Multitask learning and benchmarking with clinical time series data.
\newblock \emph{Scientific Data}, 6\penalty0 (1), Jun 2019.
\newblock ISSN 2052-4463.
\newblock \doi{10.1038/s41597-019-0103-9}.
\newblock URL \url{http://dx.doi.org/10.1038/s41597-019-0103-9}.

\bibitem[Hripcsak et~al.(2015)Hripcsak, Duke, Shah, Reich, Huser, Schuemie,
  Suchard, Park, Wong, Rijnbeek, van~der Lei, Pratt, Norén, Li, Stang,
  Madigan, and Ryan]{pmid26262116}
G.~Hripcsak, J.~D. Duke, N.~H. Shah, C.~G. Reich, V.~Huser, M.~J. Schuemie,
  M.~A. Suchard, R.~W. Park, I.~C. Wong, P.~R. Rijnbeek, J.~van~der Lei,
  N.~Pratt, G.~N. Norén, Y.~C. Li, P.~E. Stang, D.~Madigan, and P.~B. Ryan.
\newblock {{O}bservational {H}ealth {D}ata {S}ciences and {I}nformatics
  ({O}{H}{D}{S}{I}): {O}pportunities for {O}bservational {R}esearchers}.
\newblock \emph{Stud Health Technol Inform}, 216:\penalty0 574--578, 2015.

\bibitem[Johnson et~al.(2016)Johnson, Pollard, Shen, Lehman, Feng, Ghassemi,
  Moody, Szolovits, Celi, and Mark]{mimiciii}
Alistair~EW Johnson, Tom~J Pollard, Lu~Shen, Li{-}wei~H Lehman, Mengling Feng,
  Mohammad Ghassemi, Benjamin Moody, Peter Szolovits, Leo~Anthony Celi, and
  Roger~G Mark.
\newblock Mimic-iii, a freely accessible critical care database.
\newblock \emph{Scientific data}, 3:\penalty0 160035, 2016.

\bibitem[Karp(1972)]{Karp1972}
Richard~M. Karp.
\newblock \emph{Reducibility among Combinatorial Problems}, pages 85--103.
\newblock Springer US, Boston, MA, 1972.
\newblock ISBN 978-1-4684-2001-2.
\newblock \doi{10.1007/978-1-4684-2001-2_9}.
\newblock URL \url{https://doi.org/10.1007/978-1-4684-2001-2_9}.

\bibitem[{Klema} and {Laub}(1980)]{1102314}
V.~{Klema} and A.~{Laub}.
\newblock The singular value decomposition: Its computation and some
  applications.
\newblock \emph{IEEE Transactions on Automatic Control}, 25\penalty0
  (2):\penalty0 164--176, 1980.
\newblock \doi{10.1109/TAC.1980.1102314}.

\bibitem[Li et~al.(2020)Li, Rao, Solares, Hassaine, Ramakrishnan, Canoy, Zhu,
  Rahimi, and Salimi-Khorshidi]{li_behrt_2020}
Yikuan Li, Shishir Rao, José Roberto~Ayala Solares, Abdelaali Hassaine, Rema
  Ramakrishnan, Dexter Canoy, Yajie Zhu, Kazem Rahimi, and Gholamreza
  Salimi-Khorshidi.
\newblock {BEHRT}: {Transformer} for {Electronic} {Health} {Records}.
\newblock \emph{Scientific Reports}, 10\penalty0 (1):\penalty0 7155, December
  2020.
\newblock ISSN 2045-2322.
\newblock \doi{10.1038/s41598-020-62922-y}.
\newblock URL \url{http://www.nature.com/articles/s41598-020-62922-y}.

\bibitem[Luo et~al.(2020)Luo, Ye, Xiao, and Ma]{10.1145/3394486.3403107}
Junyu Luo, Muchao Ye, Cao Xiao, and Fenglong Ma.
\newblock \emph{HiTANet: Hierarchical Time-Aware Attention Networks for Risk
  Prediction on Electronic Health Records}, page 647–656.
\newblock Association for Computing Machinery, New York, NY, USA, 2020.
\newblock ISBN 9781450379984.
\newblock URL \url{https://doi.org/10.1145/3394486.3403107}.

\bibitem[Ma et~al.(2017)Ma, Chitta, Zhou, You, Sun, and
  Gao]{10.1145/3097983.3098088}
Fenglong Ma, Radha Chitta, Jing Zhou, Quanzeng You, Tong Sun, and Jing Gao.
\newblock Dipole: Diagnosis prediction in healthcare via attention-based
  bidirectional recurrent neural networks.
\newblock In \emph{Proceedings of the 23rd ACM SIGKDD International Conference
  on Knowledge Discovery and Data Mining}, KDD '17, page 1903–1911, New York,
  NY, USA, 2017. Association for Computing Machinery.
\newblock ISBN 9781450348874.
\newblock \doi{10.1145/3097983.3098088}.
\newblock URL \url{https://doi.org/10.1145/3097983.3098088}.

\bibitem[Ma et~al.(2020{\natexlab{a}})Ma, Gao, Wang, Zhang, Wang, Ruan, Tang,
  Gao, and Ma]{DBLP:conf/aaai/MaGWZWRTGM20}
Liantao Ma, Junyi Gao, Yasha Wang, Chaohe Zhang, Jiangtao Wang, Wenjie Ruan,
  Wen Tang, Xin Gao, and Xinyu Ma.
\newblock Adacare: Explainable clinical health status representation learning
  via scale-adaptive feature extraction and recalibration.
\newblock In \emph{The Thirty-Fourth {AAAI} Conference on Artificial
  Intelligence, {AAAI} 2020, The Thirty-Second Innovative Applications of
  Artificial Intelligence Conference, {IAAI} 2020, The Tenth {AAAI} Symposium
  on Educational Advances in Artificial Intelligence, {EAAI} 2020, New York,
  NY, USA, February 7-12, 2020}, pages 825--832. {AAAI} Press,
  2020{\natexlab{a}}.
\newblock URL \url{https://aaai.org/ojs/index.php/AAAI/article/view/5427}.

\bibitem[Ma et~al.(2020{\natexlab{b}})Ma, Ma, Gao, Zhang, Yu, Jiao, Ruan, Wang,
  Tang, and Wang]{ma2020covidcare}
Liantao Ma, Xinyu Ma, Junyi Gao, Chaohe Zhang, Zhihao Yu, Xianfeng Jiao, Wenjie
  Ruan, Yasha Wang, Wen Tang, and Jiangtao Wang.
\newblock Covidcare: Transferring knowledge from existing emr to emerging
  epidemic for interpretable prognosis, 2020{\natexlab{b}}.

\bibitem[Ma et~al.(2020{\natexlab{c}})Ma, Zhang, Wang, Ruan, Wang, Tang, Ma,
  Gao, and Gao]{Ma_Zhang_Wang_Ruan_Wang_Tang_Ma_Gao_Gao_2020}
Liantao Ma, Chaohe Zhang, Yasha Wang, Wenjie Ruan, Jiangtao Wang, Wen Tang,
  Xinyu Ma, Xin Gao, and Junyi Gao.
\newblock Concare: Personalized clinical feature embedding via capturing the
  healthcare context.
\newblock \emph{Proceedings of the AAAI Conference on Artificial Intelligence},
  34\penalty0 (01):\penalty0 833--840, Apr. 2020{\natexlab{c}}.
\newblock \doi{10.1609/aaai.v34i01.5428}.
\newblock URL \url{https://ojs.aaai.org/index.php/AAAI/article/view/5428}.

\bibitem[Maas et~al.(2013)Maas, Hannun, and Ng]{Maas13rectifiernonlinearities}
Andrew~L. Maas, Awni~Y. Hannun, and Andrew~Y. Ng.
\newblock Rectifier nonlinearities improve neural network acoustic models.
\newblock In \emph{in ICML Workshop on Deep Learning for Audio, Speech and
  Language Processing}, 2013.

\bibitem[Mandel et~al.(2016)Mandel, Kreda, Mandl, Kohane, and
  Ramoni]{pmid26911829}
J.~C. Mandel, D.~A. Kreda, K.~D. Mandl, I.~S. Kohane, and R.~B. Ramoni.
\newblock {{S}{M}{A}{R}{T} on {F}{H}{I}{R}: a standards-based, interoperable
  apps platform for electronic health records}.
\newblock \emph{J Am Med Inform Assoc}, 23\penalty0 (5):\penalty0 899--908, 09
  2016.

\bibitem[Mikolov et~al.(2013{\natexlab{a}})Mikolov, Chen, Corrado, and
  Dean]{mikolov2013efficient}
Tomas Mikolov, Kai Chen, Greg Corrado, and Jeffrey Dean.
\newblock Efficient estimation of word representations in vector space,
  2013{\natexlab{a}}.

\bibitem[Mikolov et~al.(2013{\natexlab{b}})Mikolov, Le, and
  Sutskever]{mikolov2013exploiting}
Tomas Mikolov, Quoc~V. Le, and Ilya Sutskever.
\newblock Exploiting similarities among languages for machine translation,
  2013{\natexlab{b}}.

\bibitem[{Nguyen} et~al.(2017){Nguyen}, {Tran}, {Wickramasinghe}, and
  {Venkatesh}]{7762861}
P.~{Nguyen}, T.~{Tran}, N.~{Wickramasinghe}, and S.~{Venkatesh}.
\newblock $\mathtt {Deepr}$: A convolutional net for medical records.
\newblock \emph{IEEE Journal of Biomedical and Health Informatics}, 21\penalty0
  (1):\penalty0 22--30, 2017.
\newblock \doi{10.1109/JBHI.2016.2633963}.

\bibitem[Paszke et~al.(2019)Paszke, Gross, Massa, Lerer, Bradbury, Chanan,
  Killeen, Lin, Gimelshein, Antiga, Desmaison, Kopf, Yang, DeVito, Raison,
  Tejani, Chilamkurthy, Steiner, Fang, Bai, and Chintala]{NEURIPS2019_9015}
Adam Paszke, Sam Gross, Francisco Massa, Adam Lerer, James Bradbury, Gregory
  Chanan, Trevor Killeen, Zeming Lin, Natalia Gimelshein, Luca Antiga, Alban
  Desmaison, Andreas Kopf, Edward Yang, Zachary DeVito, Martin Raison, Alykhan
  Tejani, Sasank Chilamkurthy, Benoit Steiner, Lu~Fang, Junjie Bai, and Soumith
  Chintala.
\newblock Pytorch: An imperative style, high-performance deep learning library.
\newblock In H.~Wallach, H.~Larochelle, A.~Beygelzimer, F.~d\textquotesingle
  Alch\'{e}-Buc, E.~Fox, and R.~Garnett, editors, \emph{Advances in Neural
  Information Processing Systems 32}, pages 8024--8035. Curran Associates,
  Inc., 2019.
\newblock URL
  \url{http://papers.neurips.cc/paper/9015-pytorch-an-imperative-style-high-performance-deep-learning-library.pdf}.

\bibitem[Pennington et~al.(2014)Pennington, Socher, and
  Manning]{pennington2014glove}
Jeffrey Pennington, Richard Socher, and Christopher~D. Manning.
\newblock Glove: Global vectors for word representation.
\newblock In \emph{Empirical Methods in Natural Language Processing (EMNLP)},
  pages 1532--1543, 2014.
\newblock URL \url{http://www.aclweb.org/anthology/D14-1162}.

\bibitem[Pollard et~al.(2018)Pollard, Johnson, Raffa, Celi, Mark, and
  Badawi]{pollard_eicu_2018}
Tom~J. Pollard, Alistair E.~W. Johnson, Jesse~D. Raffa, Leo~A. Celi, Roger~G.
  Mark, and Omar Badawi.
\newblock The {eICU} {Collaborative} {Research} {Database}, a freely available
  multi-center database for critical care research.
\newblock \emph{Scientific Data}, 5\penalty0 (1):\penalty0 180178, September
  2018.
\newblock ISSN 2052-4463.
\newblock \doi{10.1038/sdata.2018.178}.
\newblock URL \url{https://doi.org/10.1038/sdata.2018.178}.

\bibitem[Pratt(1993)]{NIPS1992_67e103b0}
L.~Y. Pratt.
\newblock Discriminability-based transfer between neural networks.
\newblock In S.~Hanson, J.~Cowan, and C.~Giles, editors, \emph{Advances in
  Neural Information Processing Systems}, volume~5, pages 204--211.
  Morgan-Kaufmann, 1993.
\newblock URL
  \url{https://proceedings.neurips.cc/paper/1992/file/67e103b0761e60683e83c559be18d40c-Paper.pdf}.

\bibitem[Rajkomar et~al.(2018)Rajkomar, Oren, Chen, Dai, Hajaj, Hardt, Liu,
  Liu, Marcus, Sun, Sundberg, Yee, Zhang, Zhang, Flores, Duggan, Irvine, Le,
  Litsch, Mossin, Tansuwan, Wang, Wexler, Wilson, Ludwig, Volchenboum, Chou,
  Pearson, Madabushi, Shah, Butte, Howell, Cui, Corrado, and
  Dean]{rajkomar_scalable_2018}
Alvin Rajkomar, Eyal Oren, Kai Chen, Andrew~M. Dai, Nissan Hajaj, Michaela
  Hardt, Peter~J. Liu, Xiaobing Liu, Jake Marcus, Mimi Sun, Patrik Sundberg,
  Hector Yee, Kun Zhang, Yi~Zhang, Gerardo Flores, Gavin~E. Duggan, Jamie
  Irvine, Quoc Le, Kurt Litsch, Alexander Mossin, Justin Tansuwan, De~Wang,
  James Wexler, Jimbo Wilson, Dana Ludwig, Samuel~L. Volchenboum, Katherine
  Chou, Michael Pearson, Srinivasan Madabushi, Nigam~H. Shah, Atul~J. Butte,
  Michael~D. Howell, Claire Cui, Greg~S. Corrado, and Jeffrey Dean.
\newblock Scalable and accurate deep learning with electronic health records.
\newblock \emph{npj Digital Medicine}, 1\penalty0 (1):\penalty0 18, December
  2018.
\newblock ISSN 2398-6352.
\newblock \doi{10.1038/s41746-018-0029-1}.
\newblock URL \url{http://www.nature.com/articles/s41746-018-0029-1}.

\bibitem[Rasmy et~al.(2021)Rasmy, Xiang, Xie, Tao, and Zhi]{PMID:34017034}
Laila Rasmy, Yang Xiang, Ziqian Xie, Cui Tao, and Degui Zhi.
\newblock Med-bert: pretrained contextualized embeddings on large-scale
  structured electronic health records for disease prediction.
\newblock \emph{NPJ digital medicine}, 4\penalty0 (1):\penalty0 86, May 2021.
\newblock ISSN 2398-6352.
\newblock \doi{10.1038/s41746-021-00455-y}.
\newblock URL \url{https://europepmc.org/articles/PMC8137882}.

\bibitem[Schönemann(1966)]{schonemann_generalized_1966}
Peter~H. Schönemann.
\newblock A generalized solution of the orthogonal procrustes problem.
\newblock \emph{Psychometrika}, 31\penalty0 (1):\penalty0 1--10, March 1966.
\newblock ISSN 1860-0980.
\newblock \doi{10.1007/BF02289451}.
\newblock URL \url{https://doi.org/10.1007/BF02289451}.

\bibitem[Shang et~al.(2019)Shang, Xiao, Ma, Li, and
  Sun]{DBLP:conf/aaai/ShangXMLS19}
Junyuan Shang, Cao Xiao, Tengfei Ma, Hongyan Li, and Jimeng Sun.
\newblock Gamenet: Graph augmented memory networks for recommending medication
  combination.
\newblock In \emph{The Thirty-Third {AAAI} Conference on Artificial
  Intelligence, {AAAI} 2019, The Thirty-First Innovative Applications of
  Artificial Intelligence Conference, {IAAI} 2019, The Ninth {AAAI} Symposium
  on Educational Advances in Artificial Intelligence, {EAAI} 2019, Honolulu,
  Hawaii, USA, January 27 - February 1, 2019}, pages 1126--1133. {AAAI} Press,
  2019.
\newblock \doi{10.1609/aaai.v33i01.33011126}.
\newblock URL \url{https://doi.org/10.1609/aaai.v33i01.33011126}.

\bibitem[S{\o}gaard et~al.(2018)S{\o}gaard, Ruder, and
  Vuli{\'c}]{sogaard-etal-2018-limitations}
Anders S{\o}gaard, Sebastian Ruder, and Ivan Vuli{\'c}.
\newblock On the limitations of unsupervised bilingual dictionary induction.
\newblock In \emph{Proceedings of the 56th Annual Meeting of the Association
  for Computational Linguistics (Volume 1: Long Papers)}, pages 778--788,
  Melbourne, Australia, July 2018. Association for Computational Linguistics.
\newblock \doi{10.18653/v1/P18-1072}.
\newblock URL \url{https://www.aclweb.org/anthology/P18-1072}.

\bibitem[Srivastava et~al.(2014)Srivastava, Hinton, Krizhevsky, Sutskever, and
  Salakhutdinov]{JMLR:v15:srivastava14a}
Nitish Srivastava, Geoffrey Hinton, Alex Krizhevsky, Ilya Sutskever, and Ruslan
  Salakhutdinov.
\newblock Dropout: A simple way to prevent neural networks from overfitting.
\newblock \emph{Journal of Machine Learning Research}, 15\penalty0
  (56):\penalty0 1929--1958, 2014.
\newblock URL \url{http://jmlr.org/papers/v15/srivastava14a.html}.

\bibitem[Steinberg et~al.(2021)Steinberg, Jung, Fries, Corbin, Pfohl, and
  Shah]{STEINBERG2021103637}
Ethan Steinberg, Ken Jung, Jason~A. Fries, Conor~K. Corbin, Stephen~R. Pfohl,
  and Nigam~H. Shah.
\newblock Language models are an effective representation learning technique
  for electronic health record data.
\newblock \emph{Journal of Biomedical Informatics}, 113:\penalty0 103637, 2021.
\newblock ISSN 1532-0464.
\newblock \doi{https://doi.org/10.1016/j.jbi.2020.103637}.
\newblock URL
  \url{https://www.sciencedirect.com/science/article/pii/S1532046420302653}.

\bibitem[Tang et~al.(2020)Tang, Davarmanesh, Song, Koutra, Sjoding, and
  Wiens]{tang_democratizing_2020}
Shengpu Tang, Parmida Davarmanesh, Yanmeng Song, Danai Koutra, Michael~W
  Sjoding, and Jenna Wiens.
\newblock Democratizing {EHR} analyses with {FIDDLE}: a flexible data- driven
  preprocessing pipeline for structured clinical data.
\newblock \emph{Journal of the American Medical Informatics Association},
  0\penalty0 (0):\penalty0 14, 2020.

\bibitem[Vaswani et~al.(2017)Vaswani, Shazeer, Parmar, Uszkoreit, Jones, Gomez,
  Kaiser, and Polosukhin]{vaswani2017attention}
Ashish Vaswani, Noam Shazeer, Niki Parmar, Jakob Uszkoreit, Llion Jones,
  Aidan~N. Gomez, Lukasz Kaiser, and Illia Polosukhin.
\newblock Attention is all you need, 2017.

\bibitem[Wojcik et~al.(2006)Wojcik, Stein, Devore, and
  Hassell]{wojcik_challenge_2006}
Barbara~E. Wojcik, Catherine~R. Stein, Raymond~B. Devore, and L.~Harrison
  Hassell.
\newblock The {Challenge} of {Mapping} between {Two} {Medical} {Coding}
  {Systems}.
\newblock \emph{Military Medicine}, 171\penalty0 (11):\penalty0 1128--1136,
  November 2006.
\newblock ISSN 0026-4075, 1930-613X.
\newblock \doi{10.7205/MILMED.171.11.1128}.
\newblock URL
  \url{https://academic.oup.com/milmed/article/171/11/1128-1136/4578127}.

\bibitem[Xiao et~al.(2018)Xiao, Choi, and Sun]{10.1093/jamia/ocy068}
Cao Xiao, Edward Choi, and Jimeng Sun.
\newblock {Opportunities and challenges in developing deep learning models
  using electronic health records data: a systematic review}.
\newblock \emph{Journal of the American Medical Informatics Association},
  25\penalty0 (10):\penalty0 1419--1428, 06 2018.
\newblock ISSN 1527-974X.
\newblock \doi{10.1093/jamia/ocy068}.
\newblock URL \url{https://doi.org/10.1093/jamia/ocy068}.

\bibitem[Xu et~al.(2021)Xu, So, and Dai]{DBLP:conf/aaai/XuSD21}
Zhen Xu, David~R. So, and Andrew~M. Dai.
\newblock {MUFASA:} multimodal fusion architecture search for electronic health
  records.
\newblock In \emph{{AAAI}}, pages 10532--10540. {AAAI} Press, 2021.

\bibitem[Zhang et~al.(2021{\natexlab{a}})Zhang, Gao, Ma, Wang, Wang, and
  Tang]{Zhang_Gao_Ma_Wang_Wang_Tang_2021}
Chaohe Zhang, Xin Gao, Liantao Ma, Yasha Wang, Jiangtao Wang, and Wen Tang.
\newblock Grasp: Generic framework for health status representation learning
  based on incorporating knowledge from similar patients.
\newblock \emph{Proceedings of the AAAI Conference on Artificial Intelligence},
  35\penalty0 (1):\penalty0 715--723, May 2021{\natexlab{a}}.
\newblock URL \url{https://ojs.aaai.org/index.php/AAAI/article/view/16152}.

\bibitem[Zhang et~al.(2021{\natexlab{b}})Zhang, Dullerud, Seyyed-Kalantari,
  Morris, Joshi, and Ghassemi]{10.1145/3450439.3451878}
Haoran Zhang, Natalie Dullerud, Laleh Seyyed-Kalantari, Quaid Morris, Shalmali
  Joshi, and Marzyeh Ghassemi.
\newblock An empirical framework for domain generalization in clinical
  settings.
\newblock In \emph{Proceedings of the Conference on Health, Inference, and
  Learning}, CHIL '21, page 279–290, New York, NY, USA, 2021{\natexlab{b}}.
  Association for Computing Machinery.
\newblock ISBN 9781450383592.
\newblock \doi{10.1145/3450439.3451878}.
\newblock URL \url{https://doi.org/10.1145/3450439.3451878}.

\end{thebibliography}

\appendix

\section{Additional justification of the motivation}
\begin{figure}[h]
\begin{center}
\includegraphics[width=\linewidth]{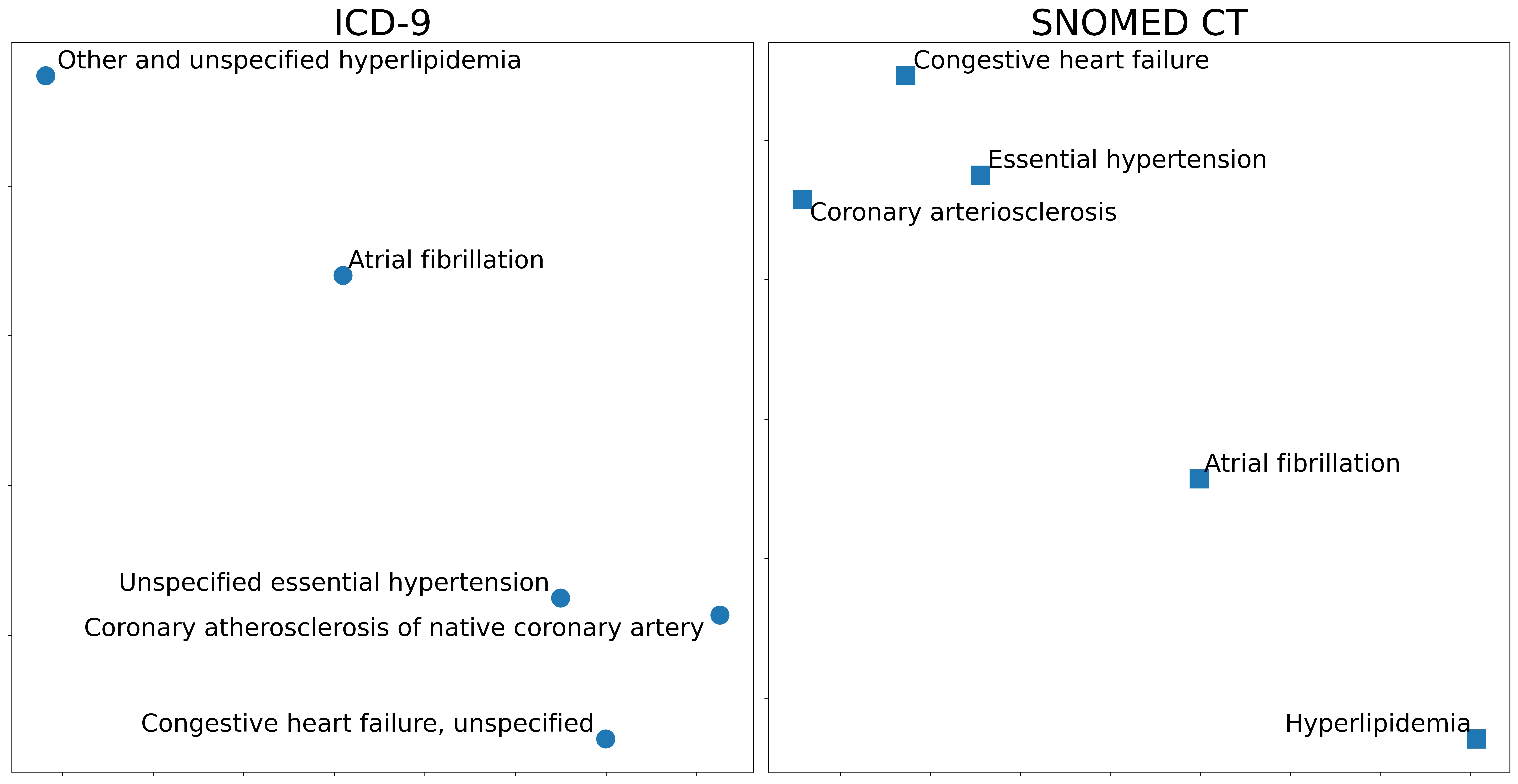}
\end{center}
\caption{Distributed medical code embedding vectors in ICD-9 (left) and SNOMED CT (right). Note that ``Coronary arteriosclerosis'' in the right figure has synonyms ``Arteriosclerotic heart disease''.} 
\label{fig:emb_space}
\end{figure}
As shown in Fig.~\ref{fig:emb_space}, we can see that these concepts have a similar structure in both spaces, suggesting that it is possible to learn an accurate mapping from one space to another. However, the task is non-trivial due to the following challenges.
\begin{itemize}[leftmargin=*]
    \item \textbf{Different granularities.} Different coding systems may have different granularities. For example, ICD-10 codes are known to be more specific than ICD-9 codes. As a result, there exist both one-to-one and one-to-many mappings.
    \item \textbf{Long-tail distribution.} The distribution of the frequencies of medical codes is often highly long-tailed. For example, in MIMIC-III~\citep{mimiciii} dataset, $50\%$ of the codes appear less than a hundred times. Consequently, the medical code embeddings are quite noisy.
\end{itemize}
To tackle these two challenges, \method learns the mapping across different coding systems in a coarse-to-fine manner. In this way, \method can better align coding systems at different granularities and alleviate the long-tail distribution of the medical codes.

\section{Notations}
\label{appendix:notations}
Notations used in this paper can be found in Tab.~\ref{tab:notations}.

\begin{table}[h]
\centering
\resizebox{\columnwidth}{!}{
\begin{tabular}{cl}
\toprule
\textbf{Notation} & \textbf{Meaning} \\
\midrule
$V, v, \mathbf{c}$ & visit sequence, single visit, medical code \\
$\mathcal{O}, \mathcal{C}, \overline{\mathcal{C}}$ & ontology, set of leaf/non-leaf codes \\
$n, m$ & \# of visits and codes \\
$\mathbf{e}, \mathbf{E}, d$ & embedding vector, matrix, and dimension \\
$\phi(\cdot)$ & code embedding mapping \\
$F(\cdot)$ & backbone model \\\
$y, \hat{y}$ & labels and predictions \\
$*S, *T$ & source and target dataset \\
\midrule
$\text{ancestor}(\mathbf{c}, l)$ & $l$-th level category for code $\mathbf{c}$ \\
$\mathbf{G}, \mathbf{D}$ & code groups and mapping dictionary \\
$\mathbf{M}, \tilde{\mathbf{M}}$ & original and sorted similarity matrix \\
$l, k$ & grouping level and \# of groups \\
$D(\cdot)$ & discriminator \\
\midrule
$\cdot, \odot$ & matrix multiplication, Hadamard product \\
$[:, :]$ & matrix indexing \\
\bottomrule
\end{tabular}}
\caption{Notations used in this paper.}
\label{tab:notations}
\end{table}

\section{Jensen-Shannon Divergence}
\label{appendix:divergence}
Jensen-Shannon divergence (JSD) can be seen as a symmetric version of Kullback–Leibler divergence (KLD), as in Eq.~\eqref{eq:jsd2},
\begin{equation}
\begin{aligned}
\text{JSD}(P \Vert Q) &= \frac{1}{2}\left( \text{KLD}(P \Vert \frac{P + Q}{2}) + \text{KLD}(Q \Vert \frac{P + Q}{2}) \right), \\
\text{KLD}(P \Vert Q) &= \underset{i}{\sum}P_i\log(\frac{P_i}{Q_i}).
\end{aligned}
\label{eq:jsd2}
\end{equation}

\section{Pseudo-Code}
\label{sec:code}

The pseudo-code of \method can be found in Algo.~\ref{alg:method}.

\begin{algorithm2e}[h]
\SetAlgoLined
\SetInd{0.5em}{0.5em}
\tcc{\textcolor{goldenbrown}{S0: Embedding}}
    Obtain target code embeddings unsupervisedly \\

\tcc{\textcolor{goldenbrown}{S1: Ontology-level Alignment}}
Initialize empty dictionary $\mathbf{D}^{(-1)}$ \\
\For{l = 0, 1, \dots, MaxDepth}{
    \tcp{\textcolor{goldenbrown}{Ontology Grouping}}
    Obtain $l$-th level coding groups $\mathbf{G}_*^{l}$ (Eq.~(\ref{eq:group},~\ref{eq:group_emb})) \\
    \tcp{\textcolor{goldenbrown}{Unsupervised Seed Induction}}
    Calculate similarity matrices $\mathbf{M}_*^{(l)}$ (Eq.~\eqref{eq:similarity_matrices}) \\
    Initialize mapping dictionary $\mathbf{D}^{(l)}$ (Eq.~\eqref{eq:init}) \\
    \tcp{\textcolor{goldenbrown}{Procrustes Optimization}}
    Merge dictionaries: $\mathbf{D}^{(l)} = \mathbf{D}^{(l)} + \mathbf{D}^{(l-1)}$ \\
    \For{number of iterative steps}{
        Find optimal $\mathbf{W}$ for current $\mathbf{D}^{(l)}$ (Eq.~\eqref{eq:update_w}) \\
        Induce new $\mathbf{D}^{(l)}$ using optimal $\mathbf{W}$ (Eq.~\eqref{eq:update_d})
    }
}
\tcc{\textcolor{goldenbrown}{Step 2: Code-level Refinement}}{
    }
    \For{number of training iterations}{
        Update discriminator $D(\cdot)$ using Eq.~\eqref{eq:loss_d} \\
        Update mapping matrix $\mathbf{W}$ using Eq.~\eqref{eq:loss_w}
    }
\caption{\method}
\label{alg:method}
\end{algorithm2e}

\section{Dataset}
\label{appendix:dataset}
We select our cohort by filtering out the following samples: (1) patients younger than 18-year-old; (2) admissions without medical codes. For MIMIC-III~\citep{mimiciii}, due to the extreme long-tail distribution of medical codes, we further filter out medical codes that appear less than 50 times in the entire dataset. We split the data into source and target sets. (For Q1-3, this is done randomly. For Q4, this is based on hospital region). Each set is then split into training, validation sets with 0.7/0.1/0.2 ratio. There is no overlap of patients between any sets. The statistics of the two datasets can be found in Tab.~\ref{tab:stat}.

\begin{table}[h!]
  \resizebox{\columnwidth}{!}{
  \begin{tabular}{l|cc}
    \toprule
    & {\bf MIMIC-III~\citep{mimiciii}} & {\bf eICU~\citep{pollard_eicu_2018}} \\
    \midrule
    \# of patients & 6,444 & 11071 \\
    Gender & M: 3,583, F: 2,861 & M: 6,209, F: 4,862 \\
   Age & 64.1 $\pm$ 16.0 & 64.3 $\pm$ 15.8 \\
   \# of visits & 17,218 & 24,997 \\
    \# of visits / patient & 2.7 $\pm$ 1.7 & 2.2 $\pm$ 0.5 \\
   \# of mortality patients & 2,460 & 1,616 \\
   Length of stay (days) & 10.7 $\pm$ 11.8 & 3.0 $\pm$ 4.8 \\
    \# of ICD-9 codes & 1,224 & 734 \\
    \# of ICD-9 codes / visit & 13.0 $\pm$ 6.4 & 4.3 $\pm$ 4.3 \\
    \# of ICD-10 codes & - & 682 \\
   \# of ICD-10 codes / visit & - & 4.2 $\pm$ 4.2 \\
    \# of NDC codes & 1,963 & - \\
   \# of NDC codes / visit & 41.1 $\pm$ 21.6 & - \\
    \bottomrule
\end{tabular}}
\caption{Statistics of the datasets. M: male. F: female.}
\label{tab:stat}
\end{table}

\section{Implementation Details}
\label{appendix:implementation_details}
For the backbone models, we set the embedding dimension to $128$. The detailed architectures are as follows:
\begin{itemize}
    \item {\bf MLP:} We first sum the code embeddings up and pass it through a linear layer with output dimension as $128$ and ReLU activation to obtain the visit embedding. The visit embedding is then fed through a linear layer to obtain the prediction scores.
    \item {\bf RNN:} We first sum the code embeddings up and pass it through the RNN layer with output dimension as $128$ to obtain the visit embedding. The visit embedding is then fed through two linear layers with hidden dimensions as $128$ and ReLU activation in between to obtain the prediction scores.
    \item {\bf RETAIN~\citep{10.5555/3157382.3157490}:} We follow the architecture described in the original work.
    \item {\bf GCT~\citep{10.1609/aaai.v34i01.5400}:} We follow the architecture described in the original work. The original model does not support multiple visits. Thus, we add an RNN layer with output dimension as $128$ at the top to model the temporal relation as suggested in the paper~\citep{10.1609/aaai.v34i01.5400}.
    \item {\bf BEHRT~\citep{li_behrt_2020}:} We follow the architecture described in the original work except that we stack 3 transformers layers with 2 attention heads as we find it works better with our dataset.
\end{itemize}

For the discriminator, we stack three linear layers with hidden dimensions as $128$, leaky rectified linear function (LeakyReLU)~\citep{Maas13rectifiernonlinearities}, and Dropout~\citep{JMLR:v15:srivastava14a} with a dropout rate $0.1$ in between.

The model is selected based on the validation set, and the performance on the test set is reported. For mortality prediction and length-of-stay estimation, we use the cross-entropy loss. We use RMSprop as the optimizer. For the teacher-student framework in step 2, we optimize the discriminator five times, followed by the mapping for one time. For all baseline methods, we apply early-stopping with patience of $20$ epochs by monitoring the validation AUC-PR for mortality prediction and AUC-ROC for length-of-stay estimation.

The tunable hyper-parameters are batch size (8, 16, 32, 64), learning rate (1e-2, 5e-3, 1e-3, 5e-4, 1e-4), and loss coefficient $\alpha$ (0.1 - 0.9) in Eq.~\eqref{eq:loss_w}. We select the hyper-parameters using random search based on the validation set performance. The final selected hyper-parameters are: batch size with 8, learning rate with 1e-4, loss coefficient $\alpha$ with 0.1.

We implement \method using PyTorch 1.6.0~\citep{NEURIPS2019_9015} and Python 3.8.5. The model is trained on an Ubuntu 20.04 machine with one AMD Ryzen Threadripper 3970X 32-Core CPU, 256GB memory, and two NVIDIA GeForce RTX 3090 GPUs.

\section{Metrics}
\label{appendix:metrics}
For mortality prediction (binary classification), we report Area Under the Receiver Operating Characteristic Curve (ROC-AUC) scores and Area Under the Precision Recall Curve (PR-AUC) scores. For length-of-stay estimation (multi-class classification), we report weighted one-v.s.-one ROC-AUC score and weighted F1 scores. 

For code mapping evaluation, we report similarity and hit@10. Similarity measures the mean cosine similarity of the transformed code embeddings between the ground truth mapping pairs. Hit@10 measures the proportion of mapped codes in the top-10 set that are correct.

For metrics ROC AUC, PR AUC, and F1, we report the average scores and standard deviation of bootstrapping for $1000$ times. For similarity and hit@10 metrics, we report the average scores and standard deviation of results from three random seeds. We perform independent two-sample t-test to evaluate if \method achieves significant improvement over baseline methods.

\section{Additional Experiments}
\label{sec: additonal_exp}
\subsection{Additional Results for Q1: Target Data with Limited Labels}
\label{appendix:q1_cont}

\begin{table}[h]
\begin{center}
\resizebox{\linewidth}{!}{
\begin{tabular}{l|l|c|c}
\toprule
\multirow{2}{*}{\bf Backbone} & \multirow{2}{*}{\bf Method} &  {\bf Mortality} & {\bf Length-of-Stay} \\
& & AUC-ROC & AUC-ROC \\
\midrule
\multirow{6}{*}{MLP} & {\it Full-Label} & {\it 0.6531 $\pm$ 0.0232} & {\it 0.2819 $\pm$ 0.0317} \\
\cline{2-4}
& Direct Training & 0.6191 $\pm$ 0.0237 & 0.5345 $\pm$ 0.0184 \\
& Transfer Learning & 0.6240 $\pm$ 0.0240 & 0.6095 $\pm$ 0.0140 \\
& MUSE & 0.6276 $\pm$ 0.0236 & 0.6240 $\pm$ 0.0157 \\
& VecMap & 0.6502 $\pm$ 0.0231 & 0.6341 $\pm$ 0.0159 \\
\rowcolor{platinum}
\cellcolor{white} & \method & {\bf 0.6631 $\pm$ 0.0228*} & {\bf 0.6350 $\pm$ 0.0160} \\
\midrule
\multirow{6}{*}{RNN} & {\it Full-Label} & {\it 0.6539 $\pm$ 0.0231} & {\it 0.2818 $\pm$ 0.0319} \\
\cline{2-4}
& Direct Training & 0.5547 $\pm$ 0.0249 & 0.4427 $\pm$ 0.0174 \\
& Transfer Learning & 0.6234 $\pm$ 0.0235 & 0.6166 $\pm$ 0.0140 \\
& MUSE & 0.6260 $\pm$ 0.0231 & 0.6367 $\pm$ 0.0159 \\
& VecMap & 0.6488 $\pm$ 0.0226 & 0.6416 $\pm$ 0.0153 \\
\rowcolor{platinum}
\cellcolor{white} & \method & {\bf 0.6627 $\pm$ 0.0223*} & {\bf 0.6487 $\pm$ 0.0161*} \\
\midrule
\multirow{6}{*}{RETAIN} & {\it Full-Label} & {\it 0.6190 $\pm$ 0.0244} & {\it 0.2648 $\pm$ 0.0302} \\
\cline{2-4}
& Direct Training & 0.5466 $\pm$ 0.0266 & 0.4427 $\pm$ 0.0174 \\
& Transfer Learning & 0.5732 $\pm$ 0.0251 & 0.5395 $\pm$ 0.0170 \\
& MUSE & 0.5838 $\pm$ 0.0252 & 0.5831 $\pm$ 0.0207 \\
& VecMap & 0.6315 $\pm$ 0.0239 & 0.5963 $\pm$ 0.0200 \\
\rowcolor{platinum}
\cellcolor{white} & \method & {\bf 0.6528 $\pm$ 0.0234*} & {\bf 0.6007 $\pm$ 0.0188*} \\
\midrule
\multirow{6}{*}{GCT} & {\it Full-Label} & {\it 0.6533 $\pm$ 0.0240} & {\it 0.2814 $\pm$ 0.0323} \\
\cline{2-4}
& Direct Training & 0.5402 $\pm$ 0.0239 & 0.4865 $\pm$ 0.0166 \\
& Transfer Learning & 0.5967 $\pm$ 0.0235 & 0.5718 $\pm$ 0.0146 \\
& MUSE & 0.6016 $\pm$ 0.0232 & 0.6129 $\pm$ 0.0153 \\
& VecMap & 0.6291 $\pm$ 0.0228 & 0.6085 $\pm$ 0.0149 \\
\rowcolor{platinum}
\cellcolor{white} & \method & {\bf 0.6539 $\pm$ 0.0229*} & {\bf 0.6363 $\pm$ 0.0147*} \\
\midrule
\multirow{6}{*}{BEHRT} & {\it Full-Label} & {\it 0.6673 $\pm$ 0.0219} & {\it 0.2652 $\pm$ 0.0275} \\
\cline{2-4}
& Direct Training & 0.5438 $\pm$ 0.0226 & 0.4730 $\pm$ 0.0175 \\
& Transfer Learning & 0.6190 $\pm$ 0.0233 & 0.5609 $\pm$ 0.0153 \\
& MUSE & 0.6040 $\pm$ 0.0227 & 0.5869 $\pm$ 0.0191 \\
& VecMap & {\bf 0.6740 $\pm$ 0.0223} & 0.6044 $\pm$ 0.0173 \\
\rowcolor{platinum}
\cellcolor{white} & \method & {0.6737 $\pm$ 0.0220} & {\bf 0.6328 $\pm$ 0.0178*} \\
\bottomrule
\end{tabular}
}
\caption{Additional results with limited labeled data in the target site (same setting as Tab.~\ref{tab:limited_label}).}
\end{center}
\end{table}

\subsection{k-Means Grouping}
\label{appendix:kmeans}
\begin{table}[h!]
\begin{center}
\resizebox{\columnwidth}{!}{
\begin{tabular}{lcc}
\toprule
\multirow{2}{*}{\bf Method} &  {\bf Mortality} & {\bf Length-of-Stay} \\
& AUC-PR & F1 \\
\midrule
\method (k-Means) & 0.2520 $\pm$ 0.0269 &  0.3605 $\pm$ 0.0179 \\
\method & {\bf 0.2712 $\pm$ 0.0280*} & {\bf 0.3744 $\pm$ 0.0182*} \\
\bottomrule
\end{tabular}}
\caption{Additional experiments in the scenario where the ontology is not available. Experiment results show that \method can adapt to the case where the medical ontology is not available.}
\label{tab:additional_exp}
\end{center}
\end{table}

We provide additional experiments in Tab.~\ref{tab:additional_exp}. \method (k-Means) uses k-Means instead of ontology to group the medical codes in step 1 (ontology-level alignment). We can see that \method (k-Means) can achieve similar performance to \method using ontology (though not as good). This shows that \method can adapt to the case where the medical ontology is not available.

\subsection{Sufficient Labeled Data Scenario}
\begin{figure}[htb]
\begin{center}
\includegraphics[width=0.75\linewidth]{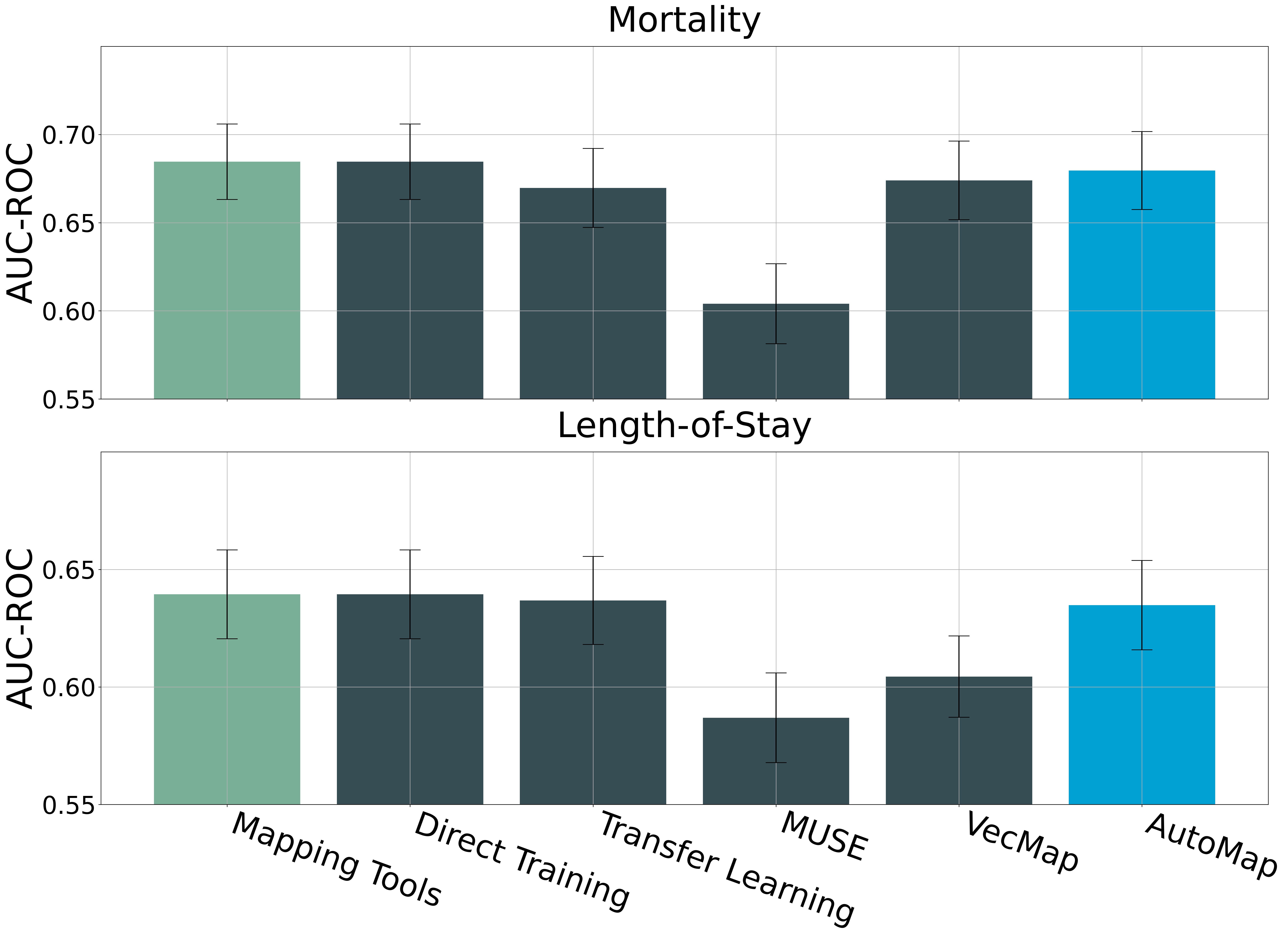}
\end{center}
\caption{Results with sufficient labeled data in the target site. Error bar denotes standard deviations. Experiment results show that \method can achieve comparable or even better performance with sufficient labeled data. Note that this is scenario is less practical in real world and is not the intended usage for \method.} 
\label{fig:exp_with_mp_and_tm}
\end{figure}

We further evaluate \method in the scenario where the target site has sufficient labeled data. We note that this scenario is less practical in the real world, as the target site may often be some small health institution with limited labeled data. The results can be found in Fig.~\ref{fig:exp_with_mp_and_tm}. In this scenario, direct training and transfer learning baselines can significantly improve since the labeled data is sufficient. Further, we can see that the code mapping tools baseline is quite strong due to the utilization of external resources. Despite these, we can see that \method still achieves comparable or even better performance with sufficient labeled data.

\end{document}